\documentclass{article} 
\usepackage{iclr2018_conference,times}
\usepackage{hyperref}
\usepackage{url}

\usepackage{booktabs}       

\usepackage{amsmath}
\usepackage{amsthm}
\usepackage{amssymb}
\usepackage{graphicx}
\usepackage{caption}
\usepackage{subcaption}
\usepackage{epstopdf}
\usepackage{wrapfig}
\usepackage{tikz}
\usepackage{xcolor}
\usetikzlibrary[arrows.meta,bending]
\usetikzlibrary{positioning}
\usepackage{algorithmic}
\usepackage{algorithm}
\usepackage{xfrac}
\usepackage{smartdiagram}
\usepackage{cancel}
\usepackage{arydshln}
\usepackage{tikz}
\usepackage{empheq}
\usetikzlibrary{arrows}
\usepackage{adjustbox}

\hypersetup{
    colorlinks,
    linkcolor={red!50!black},
    citecolor={blue!50!black},
    urlcolor={blue!80!black}
}

\newtheorem{theorem}{Theorem}

\newtheorem{definition}{Definition}
\newtheorem{remark}{Remark}

\DeclareMathOperator*{\argmin}{\arg\!\min}
\DeclareMathOperator*{\argmax}{\arg\!\max}

\DeclareMathOperator\softplus{softplus}

\newcommand{\bs}[1]{\boldsymbol{#1}} 
\renewcommand{\b}[1]{\mathbf{#1}}	
\newcommand{\dif}[1]{\mathrm{#1}} 
\newcommand{\norm}[1]{\left\lVert#1\right\rVert} 
\newcommand{\inner}[2]{\langle#1,#2\rangle} 
\newcommand{\abs}[1]{\left| #1 \right|} 
\newcommand{\parder}[2]{\frac{\partial#1}{\partial#2}} 
\newcommand{\vectorize}[1]{\text{vec}\!\left[ #1\right]} 
\newcommand{\KL}[2]{\text{KL}({#1}||{#2})} 
\newcommand{\var}[1]{\mathrm{Var}\left( #1 \right)} 
\DeclareMathOperator{\sigmoid}{sigmoid}

\def\t{\intercal} 
\def\R{\mathbb{R}}  
\def\E{\mathbb{E}}  
\def\N{\mathcal{N}} 
\def\M{\mathcal{M}} 
\def\Z{\mathcal{Z}} 
\def\X{\mathcal{X}} 
\def\C{\mathcal{C}} 
\def\Id{\mathbb{I}} 

\definecolor{minColOne}{rgb}{0, 0.4470, 0.7410}
\definecolor{minColTwo}{rgb}{0.8500, 0.3250, 0.0980}
\definecolor{minColThree}{rgb}{0.4660, 0.6740, 0.1880}

\title{Latent Space Oddity: on the Curvature \\of Deep Generative Models}

\author{Georgios Arvanitidis, Lars Kai Hansen, S{\o}ren Hauberg\\
Technical University of Denmark, Section for Cognitive Systems\\
\texttt{\{gear,lkai,sohau\}@dtu.dk}\\
}

\iclrfinalcopy 

\begin{document}

\maketitle

\begin{abstract}
Deep generative models provide a systematic way to learn nonlinear data distributions
through a set of latent variables and a nonlinear ``generator'' function that
maps latent points into the input space. The nonlinearity of the generator implies
that the latent space gives a distorted view of the input space. Under mild
conditions, we show that this distortion can be characterized by a stochastic Riemannian
metric, and we demonstrate that distances and interpolants are significantly
improved under this metric. This in turn improves probability distributions, sampling
algorithms and clustering in the latent space. Our geometric analysis further
reveals that current generators provide poor variance estimates and we propose a
new generator architecture with vastly improved variance estimates. Results are
demonstrated on convolutional and fully connected variational autoencoders, but
the formalism easily generalizes to other deep generative models.
\end{abstract}

\section{Introduction}\label{sec:intro}
  Deep generative models \citep{goodfellow:nips:2014, kingma:iclr:2014, rezende:icml:2014}
  model the data distribution of observations $\b{x} \in \X$ through corresponding
  latent variables $\b{z} \in \Z$ and a stochastic \emph{generator function} $f: \Z \rightarrow \X$ as
  \begin{align}
    \b{x} &= f(\b{z}).
    \label{eq:basic_dgm}
  \end{align}
  Using reasonably low-dimensional latent variables and highly flexible generator
  functions allows these models to efficiently represent a useful distribution
  over the underlying data manifold.
  These approaches have recently attracted a lot of attention, as deep neural
  networks are suitable generators which lead to
  the impressive performance of current \emph{variational autoencoders (VAEs)}
  \citep{kingma:iclr:2014} and \emph{generative adversarial networks (GANs)}
  \citep{goodfellow:nips:2014}.
  
\begin{figure}[t]
    \centering   
    \includegraphics[width=\textwidth]{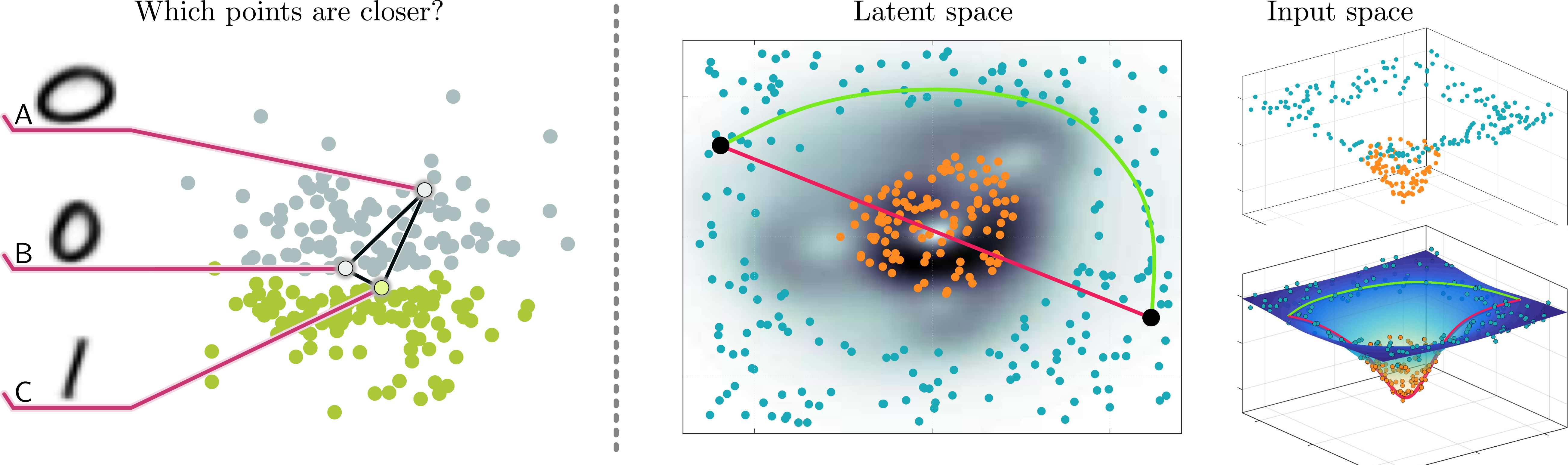} 
    \caption{\emph{Left:} An example of how latent space distances do not reflect
      actual data distances. \emph{Right:} Shortest paths on the surface spanned
      by the generator do not correspond to straight lines in the latent space,
      as is assumed by the Euclidean metric.}
    \label{fig:teaser}
  \end{figure}

  Consider the left panel of Fig.~\ref{fig:teaser}, which shows the latent representations
  of digits 0 and 1 from MNIST under a VAE. Three latent points are highlighted:
  one point (\textsf{A}) far away from the class boundary, and two points (\textsf{B}, \textsf{C})
  near the boundary, but on opposite sides. Points \textsf{B}
and \textsf{C} near the boundary seem to be very close to each other, while the third is far away from the others. Intuitively, we would hope that points from the same class
  (\textsf{A} and \textsf{B}) are closer to each other than to members of other
  classes (\textsf{C}), but this is seemingly not the case.
  \emph{In this paper, we argue this seemed conclusion is incorrect and only due
  to a misinterpretation of the latent space --- in fact points \textsf{A} and \textsf{B}
  are closer to each other than to \textsf{C} in the latent representation}.
  Correcting this misinterpretation not only improves our understanding of generative
  models, but also improves interpolations, clusterings, latent probability distributions, sampling algorithms, interpretability and more.
  
  In general, latent space distances lack physical units (making them difficult
  to interpret) and are sensitive to specifics of the underlying neural nets.
  It is therefore more robust to consider infinitesimal distances along the
  data manifold in the input space.
  Let $\b{z}$ be a latent point and let $\Delta\b{z}_1$ and $\Delta\b{z}_2$
  be infinitesimals, then we can compute the squared distance
  \begin{align}
    \norm{f(\b{z} + \Delta\b{z}_1) - f(\b{z} + \Delta\b{z}_2)}^2
      = (\Delta\b{z}_1 - \Delta\b{z}_2)^\t \left(\b{J}_{\b{z}}^\t \b{J}_{\b{z}}\right) (\Delta\b{z}_1 - \Delta\b{z}_2),
    \quad \b{J}_{\b{z}} = \frac{\partial f}{\partial \b{z}}\bigg|_{\b{z} = \b{z}},
  \end{align}
  using Taylor's Theorem.
  This implies that the natural distance function in $\Z$ changes locally as it is
  governed by the local Jacobian. Mathematically, the latent space should not then be seen as a linear Euclidean space, but rather as a curved space. The right panel of Fig.~\ref{fig:teaser} provides an example of the
  implications of this curvature. The figure shows synthetic
  data from two classes, and the corresponding latent representation
  of the data. The background color of the latent space corresponds to
  $\sqrt{\det(\b{J}_{\b{z}}^\t \b{J}_{\b{z}})}$, which can be seen as a measure of the local
  distortion of the latent space. We interpolate two points from the same class
  by walking along the connecting straight line (red); in the right panel, we
  show points along this straight line which have been mapped by the generator
  to the input space. Since the generator defines a surface in the input space,
  we can alternatively seek the shortest curve along this surface that connects the two points;
  this is perhaps the most natural choice of interpolant.
  We show this shortest curve in green.
  From the center panel it is evident that the natural interpolant is rather different
  from the straight line. This is due to the distortion of the latent space,
  which is the topic of the present paper.
  
  \paragraph{Outline.}
    In Sec.~\ref{sec:vae} we briefly present the VAE as a
    representative instance of generative models. In Sec.~\ref{sec:surfmodel} we
    connect generative models with their underlying geometry, and in
    Sec.~\ref{sec:vae_geom}  we argue that a stochastic Riemannian metric is
    naturally induced in the latent space by the generator. This metric enables us
    to compute length-minimizing curves and corresponding distances. This analysis,
    however, reveals that the traditional variance approximations in VAEs are
    rather poor and misleading; we propose a solution in Sec.~\ref{sec:proper_geometry}.
    In Sec.~\ref{sec:experiments} we demonstrate how the resulting view of the
    latent space improves latent interpolations, gives rise to more meaningful
    latent distributions, clusterings and more.
    We discuss related work in Sec.~\ref{sec:related} and conclude the paper 
    with an outlook in Sec.~\ref{sec:discuss}.
    
\section{The Variational Autoencoders acting as the Generator}\label{sec:vae}

The variational autoencoder (VAE) proposed by \citet{kingma:iclr:2014} is a simple yet powerful generative model which consists of two parts:
(1) an \emph{inference network} or \emph{recognition network} (\textit{encoder}) learns the latent representation \textit{(codes)} of the data in the input space $\X=\R^D$; and (2) the \emph{generator} (\textit{decoder}) learns how to reconstruct the data from these latent space codes in $\Z=\R^d$.

Formally, a prior distribution is defined for the latent representations
$p(\b{z}) = \N(\b{0}, \Id_d)$, and there exists a mapping function
$\bs{\mu}_{\theta}:\Z\rightarrow \X$ that generates a surface in $\X$. Moreover,
we assume that another function $\bs{\sigma}_{\theta}:\Z\rightarrow \R^D_+$ 
captures the \emph{error} (or \emph{uncertainty}) between the actual data 
observation $\b{x}\in\X$ and its reconstruction as
$\b{x} = \bs{\mu}_{\theta}(\b{z}) + \bs{\sigma}_{\theta} \odot \bs{\epsilon}$,
where $\bs{\epsilon}\sim\N(\b{0},\Id_D)$ and $\odot$ is the Hadamard (element-wise) product.
Then the likelihood is naturally defined as
$p_{\theta}(\b{x} ~|~\b{z}) = \N(\b{x} ~|~ \bs{\mu}_{\theta}(\b{z}), ~ \Id_D  \bs{\sigma}_{\theta}^2(\b{z}) )$.
The flexible functions $\bs{\mu}_{\theta}$ and $\bs{\sigma}_{\theta}$ are usually
deep neural networks with parameters $\theta$.

However, the corresponding posterior distribution $p_{\theta}(\b{z} ~|~\b{x})$ is unknown, as the marginal likelihood $p(\b{x})$ is intractable. Hence, the posterior is approximated using a variational distribution $q_{\phi}(\b{z} ~|~\b{x}) = \N(\b{z} ~|~ \bs{\mu}_{\phi}(\b{x}), ~ \Id_d \bs{\sigma}_{\phi}^2(\b{x}) )$, where the functions $\bs{\mu}_{\phi}:\X\rightarrow \Z$ and $\bs{\sigma}_{\phi}: \X \rightarrow \R^d_+$ are again deep neural networks with  parameters $\phi$. Since the generator (decoder) is a composition of linear maps and activation functions, its smoothness is based solely on the chosen activation functions.

The optimal parameters $\theta$ and $\phi$ are found by maximizing the evidence lower bound (ELBO) of the marginal likelihood $p(\b{x})$ as
\begin{align}
  \{\theta^*, \phi^*\} = \argmax_{\theta, \phi} \E_{q_{\phi}(\b{z} |\b{x})}[\log(p_{\theta}(\b{x} |\b{z}))] - \KL{q_{\phi}(\b{z} |\b{x})}{p(\b{z})},
  \label{eq:bound}
\end{align}
where the bound follows from Jensen's inequality. The optimization is based on variations of gradient descent using the reparametrization trick \citep{kingma:iclr:2014, rezende:icml:2014}. Further improvements have been proposed that provide more flexible posterior approximations \citep{rezende:icml:2015, kingma:nips:2016} or tighter lower bound \citep{burda:iclr_2016}. In this paper, we consider the standard VAE for simplicity.
The optimization problem in Eq.~\ref{eq:bound} is difficult since poor reconstructions
by $\bs{\mu}_{\theta}$ can be explained by increasing the corresponding variance
$\bs{\sigma}_{\theta}^2$. A common trick, which we also follow, is to optimize
$\bs{\mu}_{\theta}$ while keeping $\bs{\sigma}_{\theta}^2$ constant, and then
finally optimize for the variance $\bs{\sigma}_{\theta}^2$.

\section{Surfaces as the Foundation of Generative Models}\label{sec:surfmodel}
  
  Mathematically, a \emph{deterministic} generative model $\b{x} = f(\b{z})$ can be seen as a
  \emph{surface model} \citep{gauss:surfaces:1827} if the generator $f$ is sufficiently
  smooth. Here, we briefly review the basic concepts on surfaces, as they form
  the mathematical foundation of this work.
  
  \begin{wrapfigure}[12]{r}{0.41\textwidth}
    \vspace{-3mm}
    \includegraphics[width=0.41\textwidth]{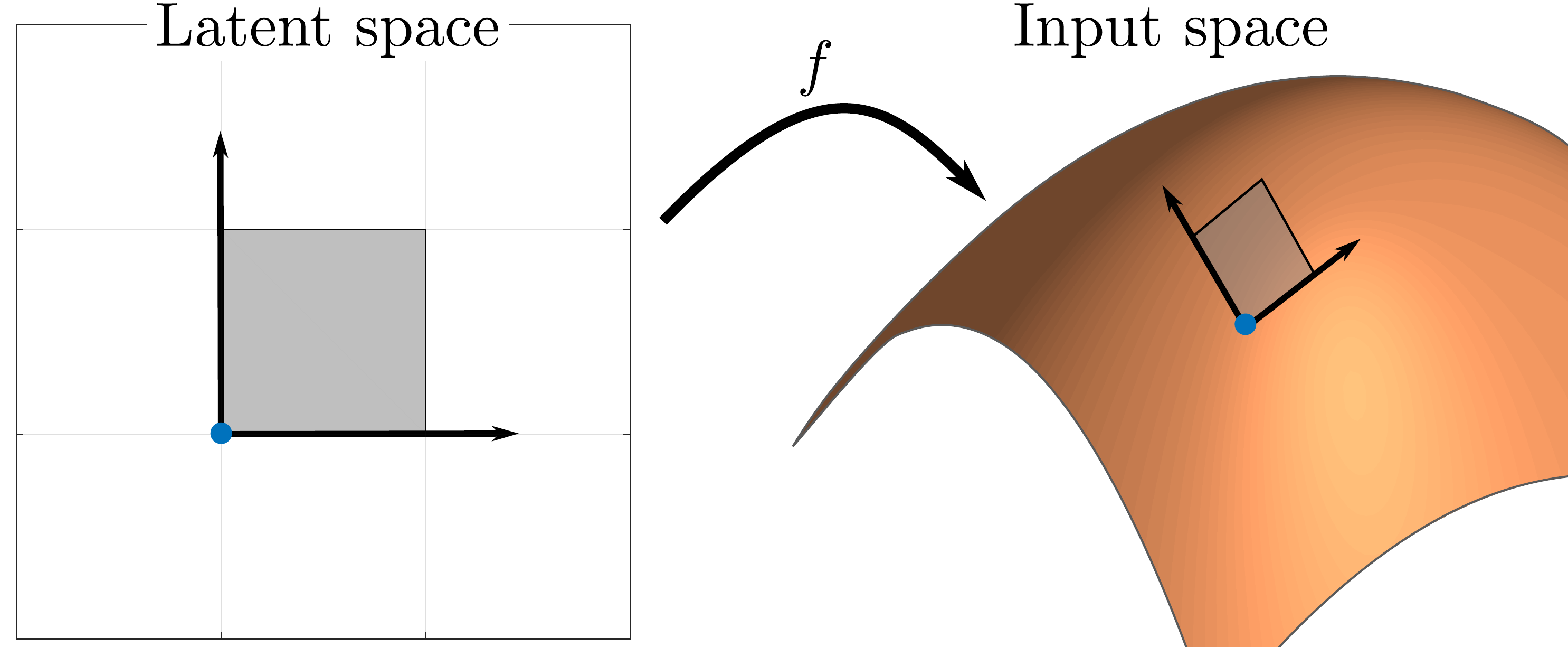}
    \vspace{-6mm}
    \caption{The Jacobian $\b{J}$ of a nonlinear function $f$ provides a local
      basis in the input space, while $\sqrt{\det(\b{J}^{\t} \b{J})}$
      measures the volume of an infinitesimal region.}
  \end{wrapfigure}
  Intuitively, a surface is a smoothly-connected set of points embedded in $\X$.
  When we want to make computations on a surface, it is often convenient to parametrize
  the surface by a low-dimensional (latent) variable $\b{z}$ along with an appropriate
  function $f: \Z \to \X$. We let $d = \mathrm{dim}(\Z)$ denote the intrinsic dimensionality of the
  surface, while $D = \mathrm{dim}(\X)$ is the dimensionality of the input space.
  If we consider a smooth (latent) curve $\bs{\gamma}_t: [0, 1] \to \Z$, then it has
  length $\int_0^1 \| \dot{\bs{\gamma}}_t \| \mathrm{d}t$, where
  $\dot{\bs{\gamma}}_t = \sfrac{\mathrm{d} \bs{\gamma}_t}{\mathrm{d} t}$
  denotes the velocity of the curve. In practice, the low-dimensional
  parametrization $\Z$ often lacks a principled meaningful metric, so we measure
  lengths in input space by mapping the curve through $f$,
  \begin{align}
    \mathrm{Length}[f(\bs{\gamma}_t)]
      &= \int_0^1 \left\| \dot{f}(\bs{\gamma}_t) \right\| \mathrm{d}t
       = \int_0^1 \left\| \b{J}_{\bs{\gamma}_t} \dot{\bs{\gamma}}_t \right\| \mathrm{d}t,
    \qquad
    \b{J}_{\bs{\gamma}_t} = \frac{\partial f}{\partial \b{z}}\bigg|_{\b{z}=\bs{\gamma}_t}
  \end{align}
  where the last step follows from Taylor's Theorem. This implies that the length
  of a curve $\bs{\gamma}_t$ along the surface can be computed directly in the latent
  space using the (locally defined) norm
  \begin{align}
    \left\| \b{J}_{\bs{\gamma}} \dot{\bs{\gamma}} \right\| 
      &= \sqrt{ (\b{J}_{\bs{\gamma}} \dot{\bs{\gamma}})^{\t} (\b{J}_{\bs{\gamma}} \dot{\bs{\gamma}}) }
       = \sqrt{ \dot{\bs{\gamma}}^{\t} (\b{J}_{\bs{\gamma}}^{\t} \b{J}_{\bs{\gamma}}) \dot{\bs{\gamma}} }
       = \sqrt{ \dot{\bs{\gamma}}^{\t} \b{M}_{\bs{\gamma}} \dot{\bs{\gamma}} }.
    \label{eq:local_norm}
  \end{align}
  Here, $\b{M}_{\bs{\gamma}} = \b{J}_{\bs{\gamma}}^{\t} \b{J}_{\bs{\gamma}}$ is a symmetric
  positive definite matrix, which acts akin to a local Mahalanobis distance measure.
  This gives rise to the definition of a \emph{Riemannian metric},
  which represents a smoothly changing inner product structure.
  \begin{definition}
    \label{def:riemannian_metric}
    A Riemannian metric $\b{M}:\Z \rightarrow \R^{d\times d}$ is a smooth function
    that assigns a symmetric positive definite matrix to any point in $\Z$.
  \end{definition}
  It should be clear that if the generator function $f$ is sufficiently smooth,
  then $\b{M}_{\bs{\gamma}}$ in Eq.~\ref{eq:local_norm} is a Riemannian metric.
  
  When defining distances across a given surface, it is meaningful to seek
  the shortest curve connecting two points. Then a distance can be defined as
  the length of this curve. The shortest curve connecting points $\b{z}_0$
  and $\b{z}_1$ is by (trivial) definition
  \begin{align}
    \bs{\gamma}_t^{\text{(shortest)}}
      &= \argmin_{\bs{\gamma}_t} \mathrm{Length}[f(\bs{\gamma}_t)],
    \qquad
    \bs{\gamma}_0 = \b{z}_0, \enspace \bs{\gamma}_1 = \b{z}_1.
  \end{align}
  A classic result of differential geometry \citep{docarmo:1992} is that
  solutions to this optimization problem satisfy the following system of
  ordinary differential equations (ODEs)
  \begin{align}
    \ddot{\bs{\gamma}}_t 
      &=-0.5\cdot\b{M}_{\bs{\gamma}_t}^{-1}\left[2\cdot (\dot{\bs{\gamma}}_t^\t \otimes \Id_d) \partial_{\bs{\gamma}_t}{\vectorize{\b{M}_{\bs{\gamma}_t}}}\dot{\bs{\gamma}}_t - \partial_{\bs{\gamma}_t}{\vectorize{\b{M}_{\bs{\gamma}_t}}}^\t (\dot{\bs{\gamma}}_t \otimes \dot{\bs{\gamma}}_t)\right]\label{eq:ode}\\
      &= -0.5\cdot\b{M}_{\bs{\gamma}_t}^{-1}\Big[2\cdot\partial_{\bs{\gamma}_t}\b{M}_{\bs{\gamma}_t} - \partial_{\bs{\gamma}_t}{\vectorize{\b{M}_{\bs{\gamma}_t}}}^\t \Big] (\dot{\bs{\gamma}}_t \otimes \dot{\bs{\gamma}}_t)\label{eq:ode_new},
  \end{align}
  where $\partial_{\bs{\gamma}_t}\b{M}_{\bs{\gamma}_t}=\left[\partial_{\gamma^{(1)}_t} \b{M}_{\bs{\gamma}_t},\dots, \partial_{\gamma^{(d)}_t} \b{M}_{\bs{\gamma}_t}\right]$ with $\partial_{\gamma^{(j)}_t} \b{M}_{\bs{\gamma}_t}\in\R^{d\times d}$ the partial derivative of $\b{M}_{\bs{\gamma}_t}$ with respect to the $j$-th component of the curve, $\vectorize{\cdot}$ stacks the columns of a matrix into a vector so $\partial_{\bs{\gamma}_t}\vectorize{\b{M}_{\bs{\gamma}_t}}\in\R^{d^2 \times d}$ and
  $\otimes$ is the Kronecker product. Note that Eq.~\ref{eq:ode_new} enables us to compute the associated Christoffel symbols. For completeness, we provide the derivation of the ODEs in Appendix~\ref{sec:geodesic_system_derivation}, as well as the steps for computing the Christoffel symbols.
  Shortest curves can then be computed by solving the ODEs numerically; our implementation
  uses \texttt{bvp5c} from Matlab.

\section{The Geometry of Stochastic Generators}\label{sec:vae_geom}

  In the previous section, we considered \emph{deterministic} generators $f$ to
  provide relevant background information. We now extend these results to the
  stochastic case; in particular we consider
  \begin{align}
    f(\b{z}) = \bs{\mu}(\b{z}) + \bs{\sigma}(\b{z}) \odot \bs{\epsilon},
    \qquad
    \bs{\mu}: \Z \to \X,\enspace
    \bs{\sigma}: \Z \to \R_{+}^D,\enspace
    \bs{\epsilon} \sim \N(\b{0},\Id_D).
    \label{eq:stoch_gen}
  \end{align}
  This is the generator driving VAEs and related models.
  For our purposes, we will call $\bs{\mu}(\cdot)$ the \emph{mean function} and
  $\bs{\sigma}^2(\cdot)$ the \emph{variance function}.
  
  Following the discussion from the previous section, it is natural to consider
  the Riemannian metric $\b{M}_{\b{z}} = \b{J}_{\b{z}}^{\t}\b{J}_{\b{z}}$ in the latent space. Since
  the generator is now stochastic, this metric also becomes stochastic, which
  complicates analysis. The following results, however, simplify matters.
  \begin{theorem}
  \label{thm:expected_metric}
    If the stochastic generator in Eq.~\ref{eq:stoch_gen} has mean and variance
    functions that are at least twice differentiable, then the expected metric
    equals
    \begin{align}
      \overline{\b{M}}_{\b{z}} =
      \E_{p(\bs{\epsilon})}[ \b{M}_{\b{z}} ] 
        = \left( \b{J}^{(\bs{\mu})}_{\b{z}} \right)^\t \left( \b{J}^{(\bs{\mu})}_{\b{z}} \right)
        + \left( \b{J}^{(\bs{\sigma})}_{\b{z}} \right)^\t \left( \b{J}^{(\bs{\sigma})}_{\b{z}} \right),
    \end{align}
    where $\b{J}^{(\bs{\mu})}_{\b{z}}$ and $\b{J}^{(\bs{\sigma})}_{\b{z}}$
    are the Jacobian matrices of $\bs{\mu}(\cdot)$ and $\bs{\sigma}(\cdot)$.
  \end{theorem}
  %
  %
  \begin{remark}
    By Definition~\ref{def:riemannian_metric}, the metric tensor must
    change smoothly, which implies that the Jacobians must be smooth functions as well.
    This is easily ensured with activation functions for the neural networks
    that are $\C^2$ differentiable, e.g.\ $\tanh(\cdot$), $\sigmoid(\cdot)$, and $\softplus(\cdot)$.
  \end{remark}
  \begin{theorem}[Due to \citet{Tosi:UAI:2014}]\label{theom:var}
    The variance of the metric under the $L_2$ measure vanishes when the data dimension
    goes to infinity, i.e.\ 
      $\lim_{D\to\infty}\var{\b{M}_{\b{z}}} = 0$.
  \end{theorem}
  \begin{wrapfigure}[11]{r}{0.3\textwidth}
    \vspace{-3mm}
    \includegraphics[width=0.3\textwidth]{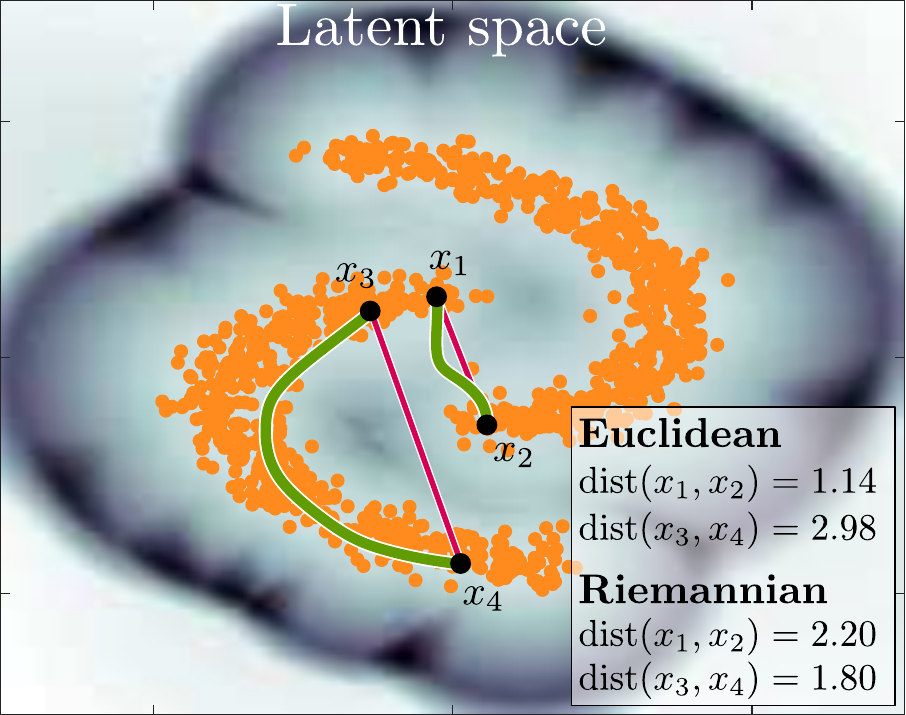}
    \vspace{-6mm}
    \caption{Example shortest paths and distances.}
      \label{fig:example_interpolation}
  \end{wrapfigure}
  Theorem~\ref{theom:var} suggests that the (deterministic) expected metric
  $\overline{\b{M}}_{\b{z}}$ is a good approximation to the underlying stochastic
  metric when the data dimension is large. We make this approximation, which allows
  us to apply the theory of deterministic generators. 
  
  This expected metric has a particularly appealing form, where the
  two terms capture the distortion of the mean and the variance functions
  respectively. In particular, the variance term
  $(\b{J}^{(\bs{\sigma})}_{\b{z}})^\t (\b{J}^{(\bs{\sigma})}_{\b{z}})$
  will be large in regions of the latent space, where the generator has
  large variance. This implies that induced distances will be large in regions
  of the latent space where the generator is highly uncertain, such that
  shortest paths will tend to avoid these regions. These paths will then tend to
  follow the data in the latent space, c.f.\ Fig.~\ref{fig:example_interpolation}. It is worth stressing, that no learning 
  is needed to compute this metric: it only consists of terms that can be derived directly through $f$. 


\subsection{Ensuring Proper Geometry Through Meaningful Variance Functions}\label{sec:proper_geometry}
  Theorem~\ref{thm:expected_metric} informs us about how the geometry of the generative
  model depends on both the mean and the variance of the generator. Assuming
  successful training of the generator, we can expect to have good estimates
  of the geometry in regions near the data. But what happens in regions further
  away from the data? In general, the mean function cannot be expected to give
  useful extrapolations to such regions, so it is reasonable to require that
  the generator has high variance in regions that are not near the data.
  In practice, the neural net used to represent the variance function is only trained
  in regions where data is available, which implies that variance estimates are
  extrapolated to regions with no data. As neural nets tend to extrapolate poorly,
  \emph{practical variance estimates tend to be arbitrarily poor in regions without data}.

  \begin{figure}[ht!]
    \centering
    \begin{subfigure}[b]{0.235\textwidth}
      \centering
      \tiny{Input space}\\
      \includegraphics[height=2.5cm]{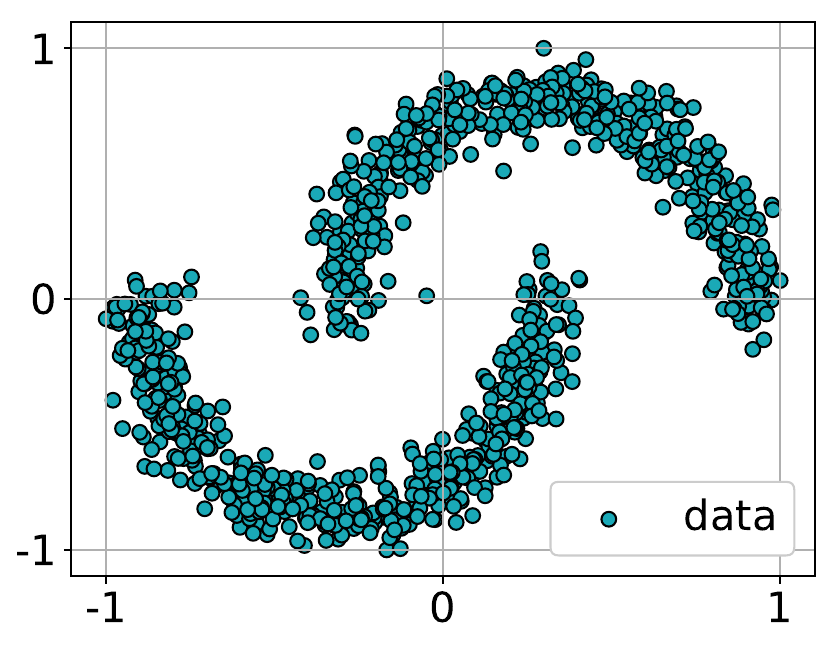}
    \end{subfigure}
    ~
    \begin{subfigure}[b]{0.235\textwidth}
      \centering
      \tiny{Latent space}\\
      \includegraphics[height=2.5cm]{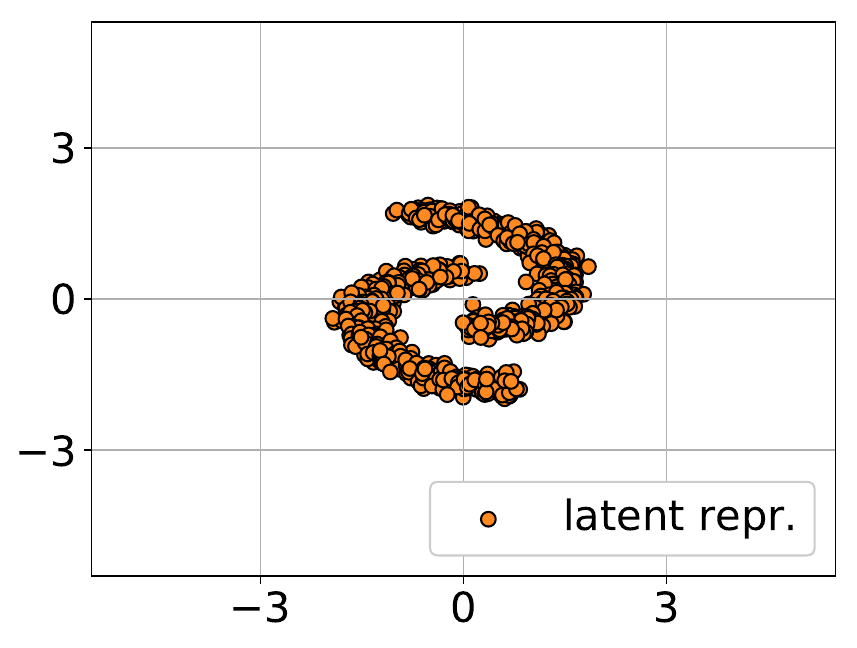}
    \end{subfigure}
    ~ 
    \begin{subfigure}[b]{0.235\textwidth}
      \centering
      \tiny{Standard variance estimate}\\
      \includegraphics[height=2.5cm]{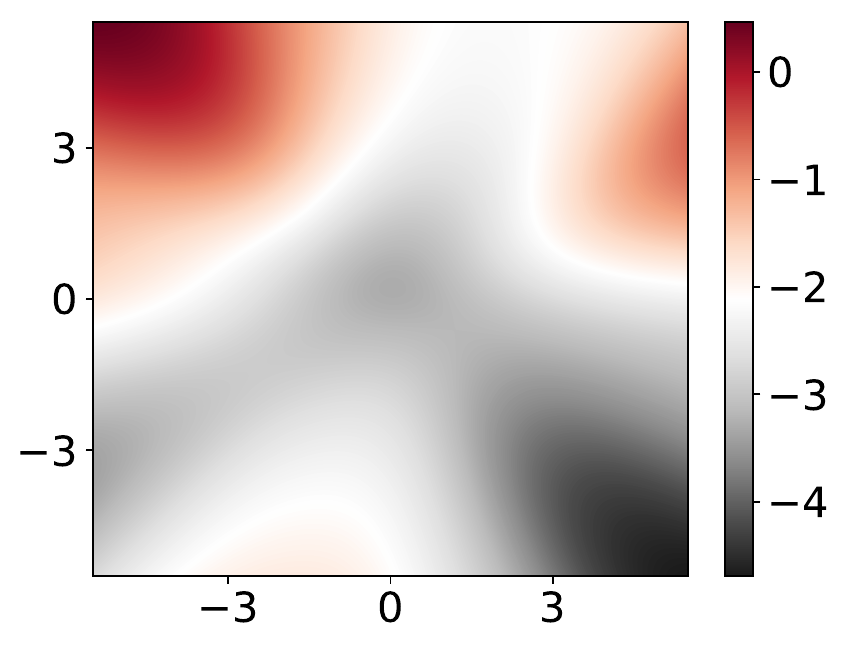}
    \end{subfigure}
    ~ 
    \begin{subfigure}[b]{0.235\textwidth}
      \centering
      \tiny{Proposed variance estimate}\\
      \includegraphics[height=2.5cm]{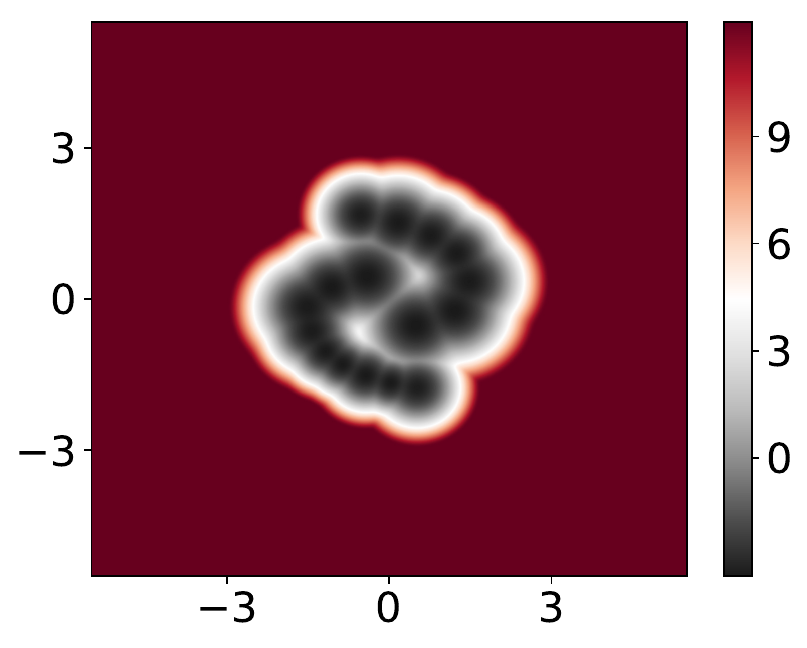}
    \end{subfigure}
    \caption{From left to right: training data in $\X$, latent representations in $\Z$,
      the standard deviation $\log(\sum_{j=1}^D \sigma_j(\b{z}))$ for the standard
      variance network, and the proposed solution.}
    \label{fig:geometry_problem}
  \end{figure}

  Figure~\ref{fig:geometry_problem} illustrates this problem. The first two panels
  show the data and its corresponding latent representations (here both input
  and latent dimensions are 2 to ease illustration). The third panel shows the
  variance function under a standard architecture, deep multilayer perceptron with \textit{softplus} nonlinearity for the output layer.
  It is evident that variance estimates in regions without data are not representative
  of either uncertainty or error of the generative process; sometimes variance
  is high, sometimes it is low. From a probabilistic modeling point-of-view, this
  is disheartening. An informal survey of publicly available VAE implementations
  also reveals that it is common to enforce a constant unit variance everywhere;
  this is further disheartening.
  
  For our purposes, we need well-behaved variance functions to ensure a well-behaved
  geometry, but reasonable variance estimates are of general use. Here, 
  as a general strategy, we propose to model the inverse variance with a network that
  extrapolates towards zero. This at least ensures that variances are large
  in regions without data. Specifically, we model the \emph{precision} as 
  $\bs{\beta}_{\psi}(\b{z}) = \frac{1}{\bs{\sigma}_{\psi}^2(\b{z})}$, where all
  operations are element-wise. Then, we model this precision with a 
  \emph{radial basis function (RBF) neural network} \citep{que:aistats:2016}.
  Formally this is written 
  \begin{align}
    \bs{\beta}_{\psi}(\b{z})
      &= {\b{W}\b{v}(\b{z})} + \bs{\zeta}, \quad \text{with} \quad v_k(\b{z}) = \exp\left(-\lambda_k \norm{\b{z}-\b{c}_k}_2^2\right),
         ~k=1,\dots,K,
  \end{align}
  where $\psi$ are all parameters, $\b{W} \in \R^{D\times K}_{> 0}$ are the 
  \emph{positive} weights of the network (positivity ensures a positive precision),
  $\b{c}_k$ and $\lambda_k$ are the centers and the bandwidth of the $K$ radial basis
  functions, and $\bs{\zeta}\rightarrow \b{0}$ is a
  vector of positive constants to prevent division by zero. It is easy to see
  that with this approach the variance of the generator increases with the distance
  to the centers. The right-most panel of Fig.~\ref{fig:geometry_problem} shows
  an estimated variance function, which indeed has the desired property that variance
  is large outside the data support. Further, note the increased variance between
  the two clusters, which captures that even interpolating between clusters
  comes with a level of uncertainty. 
  In Appendix~\ref{sec:marginal_likelihood_modeling} we also demonstrate that this simple
  variance model improves the marginal likelihood $p(\b{x})$ on held-out data.

  \begin{wrapfigure}[17]{r}{0.33\textwidth}
    %
    \vspace{-9mm}
    \includegraphics[width=0.31\textwidth]{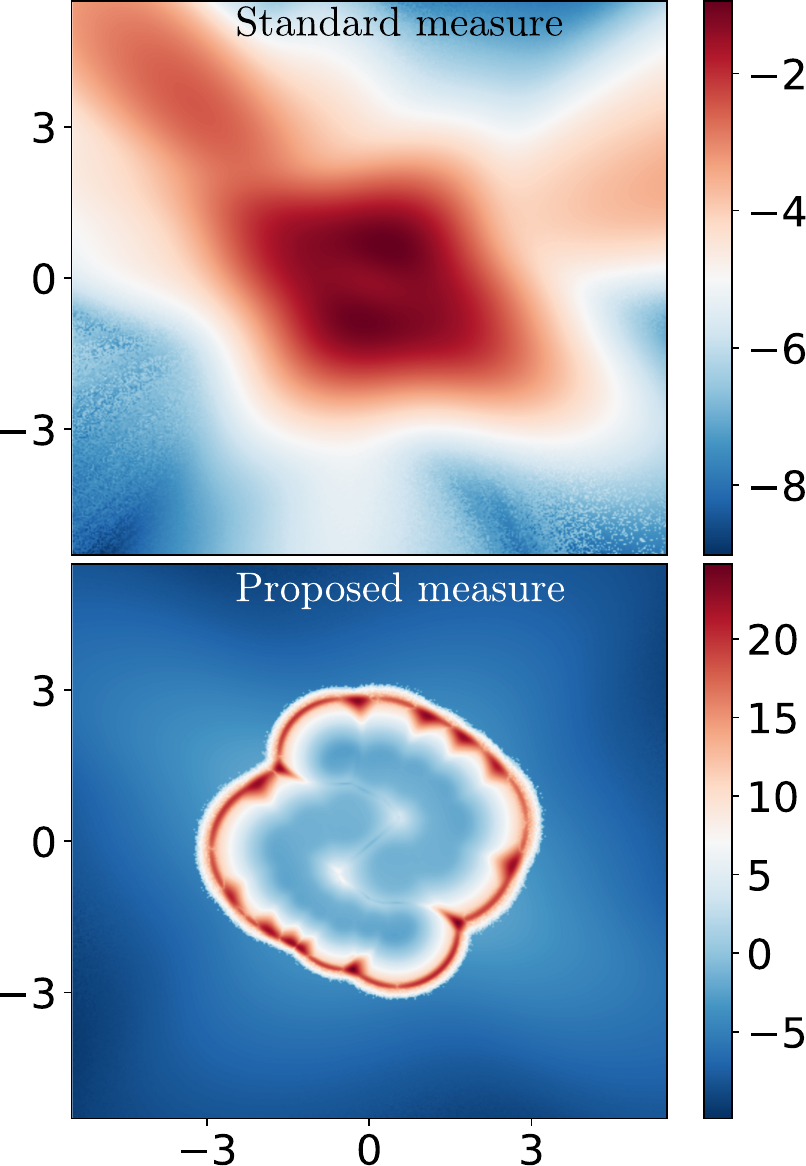}
    \vspace{-2mm}
    \caption{Comparison of (log) measures of standard (top) and proposed (bottom) variances.}
      \label{fig:measure_comparison}
  \end{wrapfigure}
  Training the variance network amounts to fitting the RBF network. Assuming we
  have already trained the inference network (Sec.~\ref{sec:vae}), we can encode
  the training data, and use $k$-means to estimate the RBF centers. Then, an
  estimate for the bandwidths of each kernel can be computed as
  \begin{align}
    \label{eq:rbf_bandwidth}
    \lambda_k = \frac{1}{2} \left(a \frac{1}{\abs{\C_k}}\sum_{\b{z}_j \in \C_k} \norm{\b{z}_j - \b{c}_k}_2 \right)^{-2}
  \end{align}
  where the hyper-parameter $a\in\R_+$ controls the curvature of the Riemannian
  metric, i.e. how fast it changes based on the uncertainty. Since the mean
  function of the generator is already trained, the weights of the RBF can be
  found using projected gradient descent to ensure positive weights.

  One visualization of the distortion of the latent space relative to the input
  space is the geometric \emph{volume measure} $\sqrt{\det(\b{M}_{\b{z}})}$,
  which captures the volume of an infinitesimal area in the input space.
  Figure~\ref{fig:measure_comparison} shows this volume measure for both standard
  variance functions as well as our proposed RBF model. We see that the 
  proposed model captures the trend of the data, unlike the standard model.

\section{Empirical Results}\label{sec:experiments}
  We demonstrate the usefulness of the geometric view of the latent space with
  several experiments. Model and implementation details can be found in Appendix~\ref{sec:model_details}.
  In all experiments we first train a VAE and then use the induced Riemannian
  metric.

\subsection{Meaningful Distances}\label{ex:algorithm_extension}

\begin{figure}[h]
    \centering    
    \begin{subfigure}[b]{0.31\textwidth}
        \includegraphics[width=\textwidth]{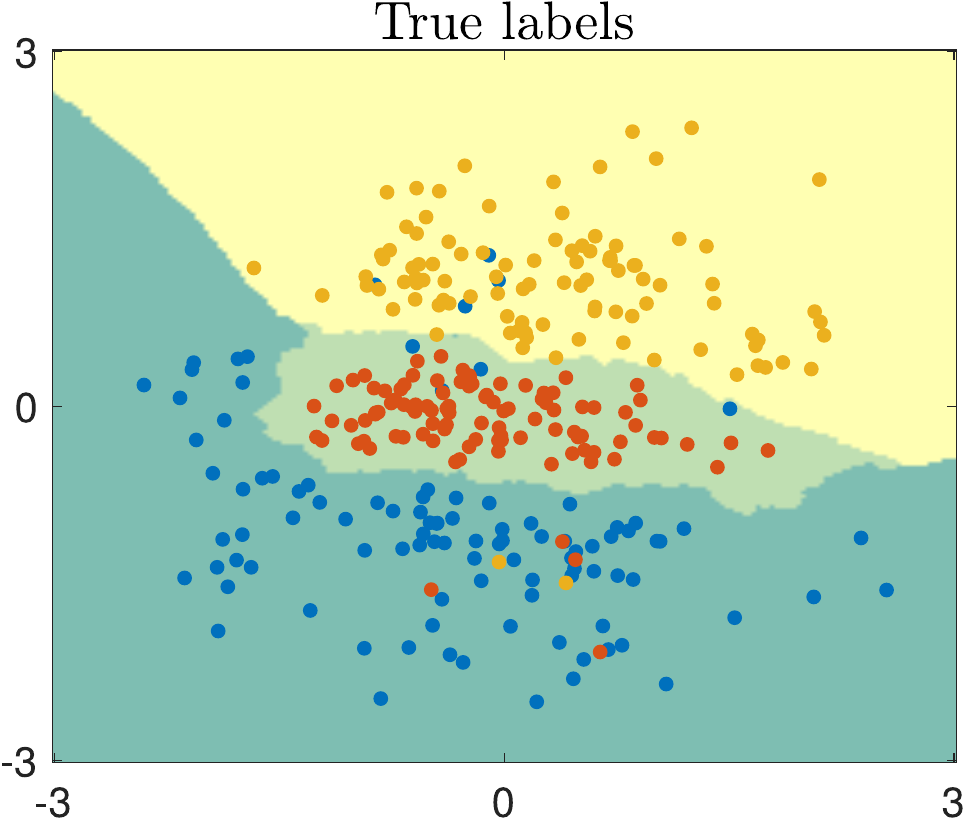}
    \end{subfigure}
    ~
    \begin{subfigure}[b]{0.31\textwidth}
        \includegraphics[width=\textwidth]{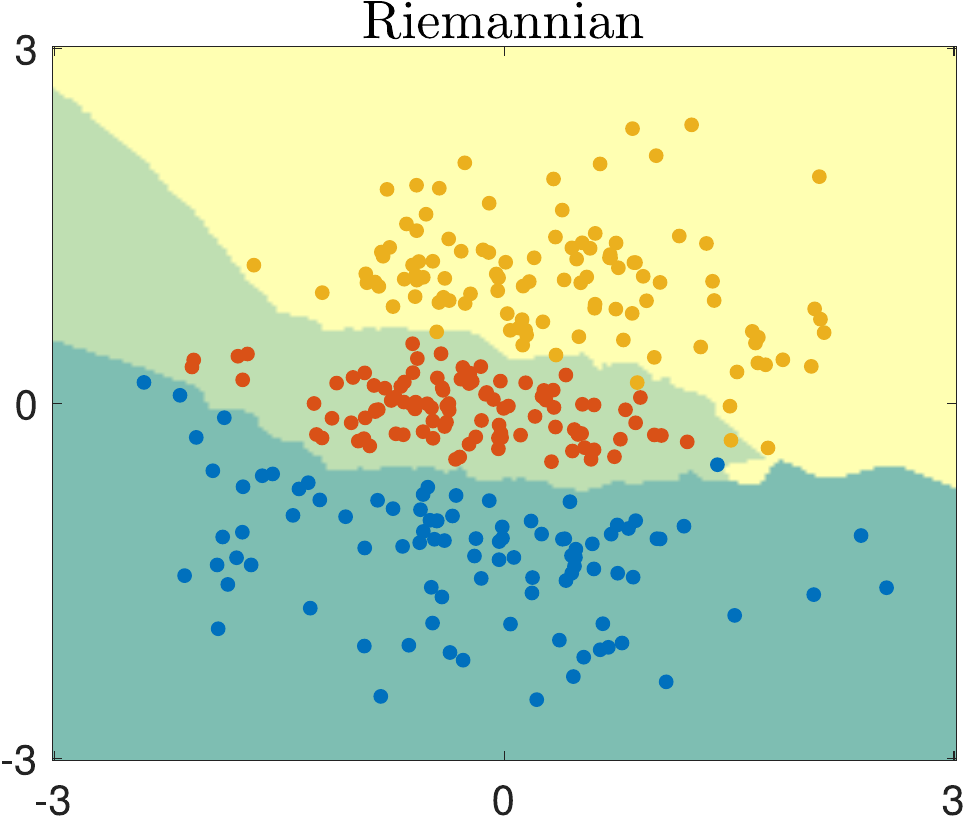}
    \end{subfigure}
    ~
    \begin{subfigure}[b]{0.31\textwidth}
        \includegraphics[width=\textwidth]{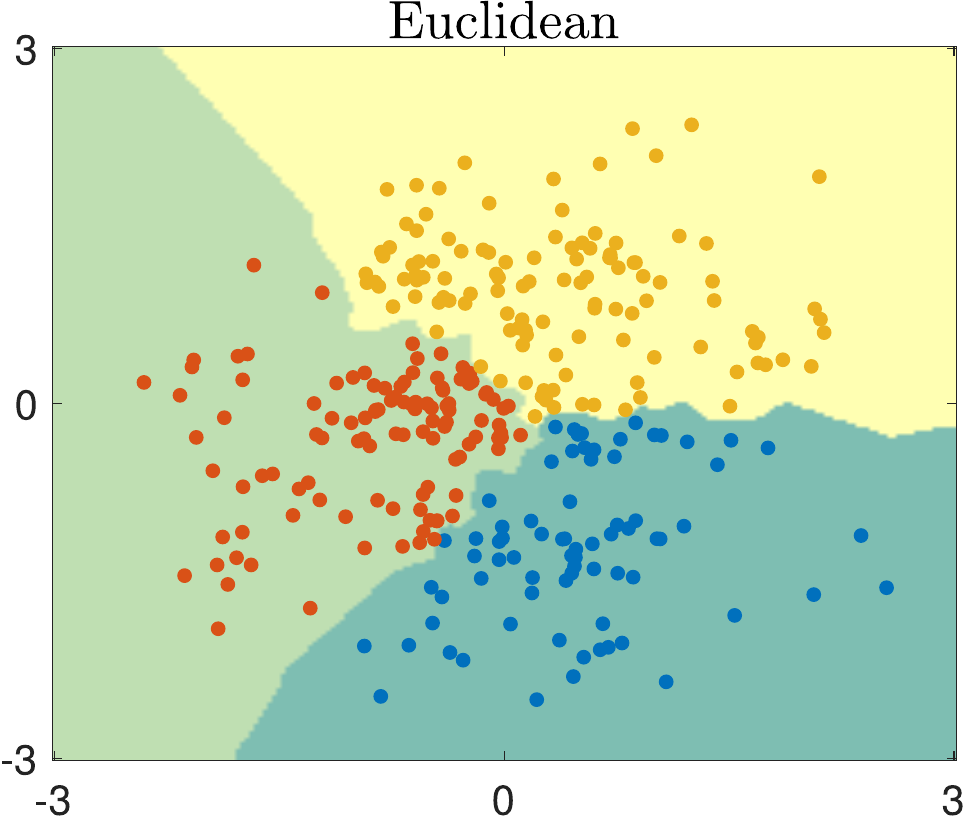}
    \end{subfigure}
	\caption{The result of $k$-means comparing the distance measures. For the 
	  decision boundaries we used 7-NN classification.}
    \label{fig:deep_kmeans}
  \end{figure}

  \begin{wraptable}{r}{7.5cm}
  \vspace{-16pt}
  \centering
  \begin{tabular}{ c c c  }
    \toprule
    Digits & Linear & Riemannian \\ 
    \midrule
    $\{0,1,2\}$ & $77.57 (\pm 0.87)\% $ & $\mathbf{94.28} (\pm \mathbf{1.14})\% $ \\  
    $\{3,4,7\}$ & $77.80 (\pm 0.91)\% $ & $\mathbf{89.54} (\pm \mathbf{1.61})\% $ \\
    $\{5,6,9\}$ & $64.93 (\pm 0.96)\% $ & $\mathbf{81.13} (\pm \mathbf{2.52})\% $ \\
    \bottomrule
  \end{tabular}
  \caption{The $F$-measure results for $k$-means.}
  \label{table:deep_kmeans}
  \vspace{-10pt}
  \end{wraptable}
  First we seek to quantify if the induced Riemannian distance in the latent
  space is more useful than the usual Euclidean distance. For this we
  perform basic $k$-means clustering under the two metrics.
  We construct 3 sets of MNIST digits, using 1000 random samples for each digit.
  We train a VAE for each set, and then subdivide each into 10 sub-sets, and
  performed $k$-means clustering under both distances. One example result
  is shown in Fig.~\ref{fig:deep_kmeans}. Here it is evident that, since the latent
  points roughly follow a unit Gaussian, there is little structure to be discovered
  by the Euclidean $k$-means, and consequently it performs poorly. The Riemannian
  clustering is remarkably accurate.
  Summary statistics across all subsets are provided in Table~\ref{table:deep_kmeans},
  which shows the established $F$-measure for clustering accuracy. Again, the Riemannian
  metric significantly improves clustering. This implies that the underlying
  Riemannian distance is more useful than its Euclidean counterpart.

\subsection{Interpolations}\label{sec:interpolate}
\begin{figure}[h]
    \centering

  	\includegraphics[width=0.8\textwidth]{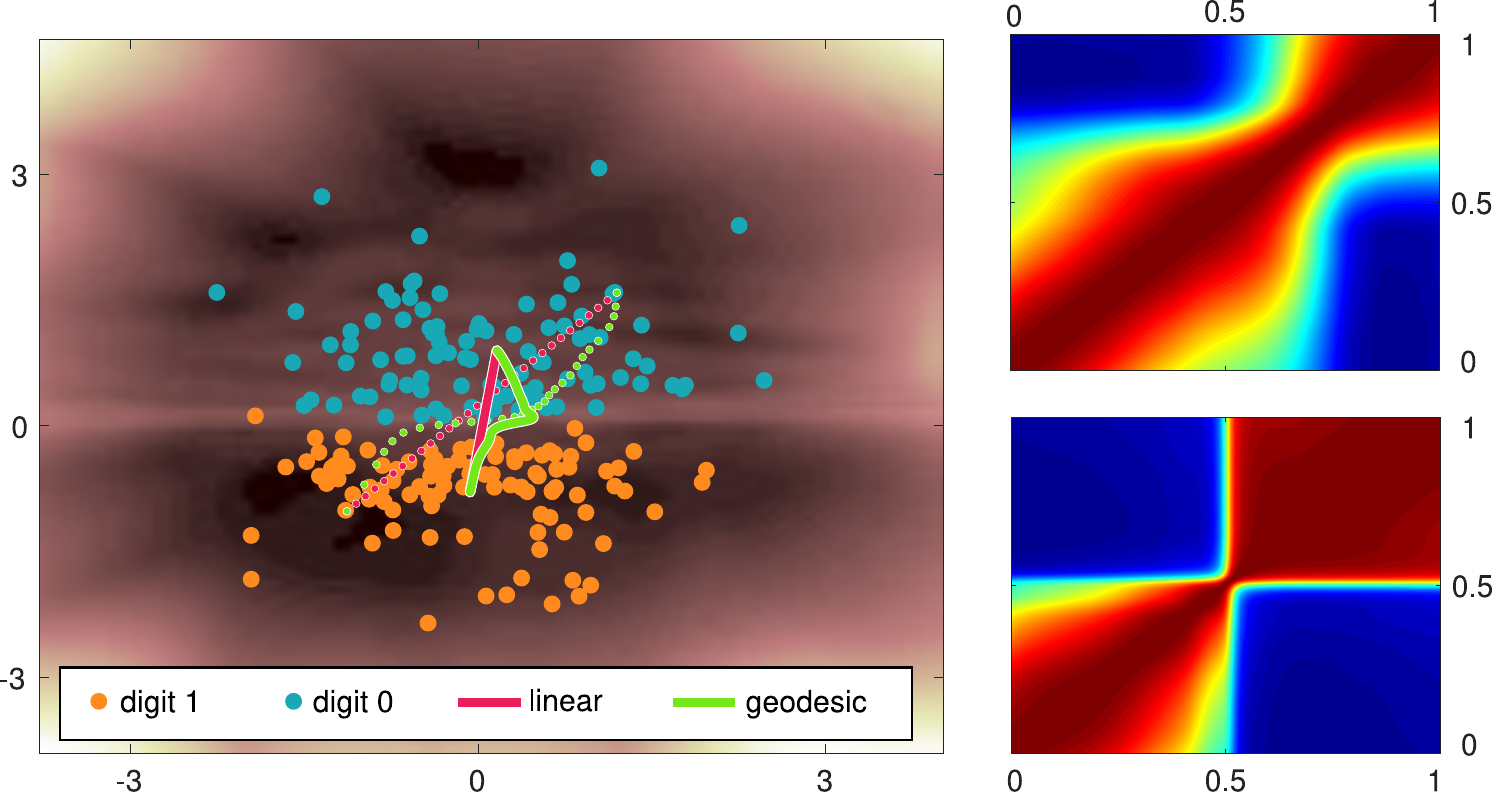} 
	\includegraphics[width=1\textwidth, height=4cm]{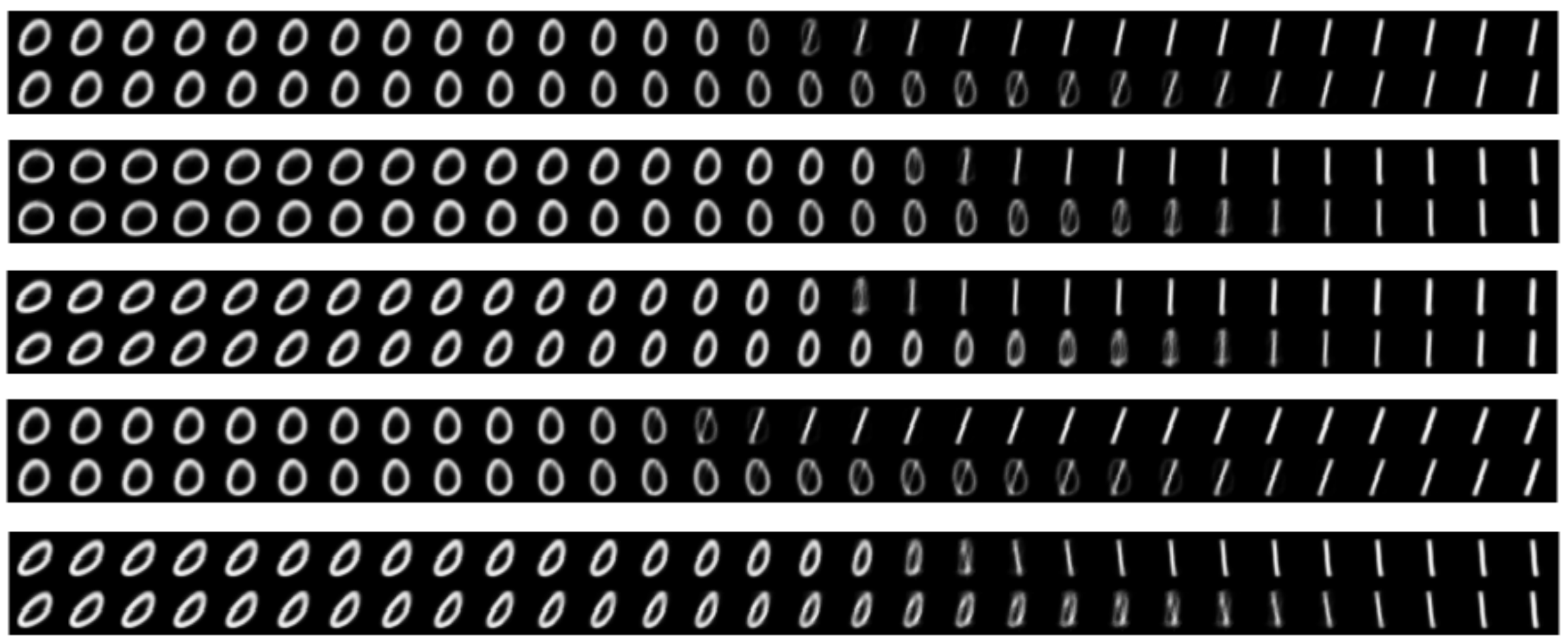}
    \caption{\emph{Left:} the latent space with example interpolants.
      \emph{Right:} auto-correlations of Riemannian (top) and Euclidean (bottom)
       samples along the curves of the left panel.
      \emph{Bottom:} decoded images along Euclidean (top rows) and Riemannian (bottom rows) interpolants.
      }
        \label{fig:deep_land_interpolations}
  \end{figure}

\begin{figure}[h]
\centering
  	\includegraphics[width=1\textwidth]{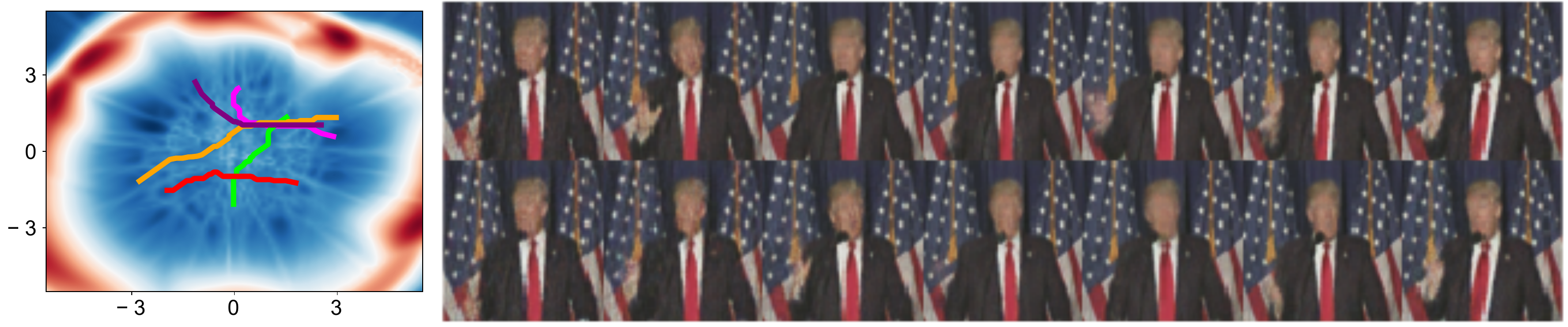} 
  	\caption{\textit{Left:} the latent space and geodesic interpolants. \textit{Right:} samples comparing Euclidean (top row) with Riemannian (bottom row) interpolation. Corresponding videos can be found \href{https://www.dropbox.com/sh/8ncrsmhaf3x1s80/AAD-JBY-5PcZGQPzXilreSOaa?dl=0}{here}.}
  	\label{fig:trump_interpolations}
\end{figure}

  Next, we investigate whether the Riemannian metric gives more meaningful interpolations.
  First, we train a VAE for the digits 0 and 1 from MNIST. The upper left panel
  of Fig.~\ref{fig:deep_land_interpolations} shows the latent space with the
  Riemannian measure as background color, together with two interpolations.
  Images generated by both Riemannian and Euclidean interpolations are shown in
  the bottom of Fig.~\ref{fig:deep_land_interpolations}. The Euclidean
  interpolations seem to have a very abrupt change when transitioning from one
  class to another. The Riemannian interpolant gives smoother changes in the
  generated images. The top-right panel of the figure shows the auto-correlation
  of images along the interpolants; again we see a very abrupt change in the
  Euclidean interpolant, while the Riemannian is significantly smoother.
  We also train a convolutional VAE on frames from a video.
  Figure~\ref{fig:trump_interpolations} shows the corresponding latent space
  and some sample interpolations. As before, we see more smooth changes
  in generated images when we take the Riemannian metric into account.

\subsection{Latent Probability Distributions}\label{ex:land}
\begin{figure}[h]
    \centering    
    \includegraphics[width=\textwidth]{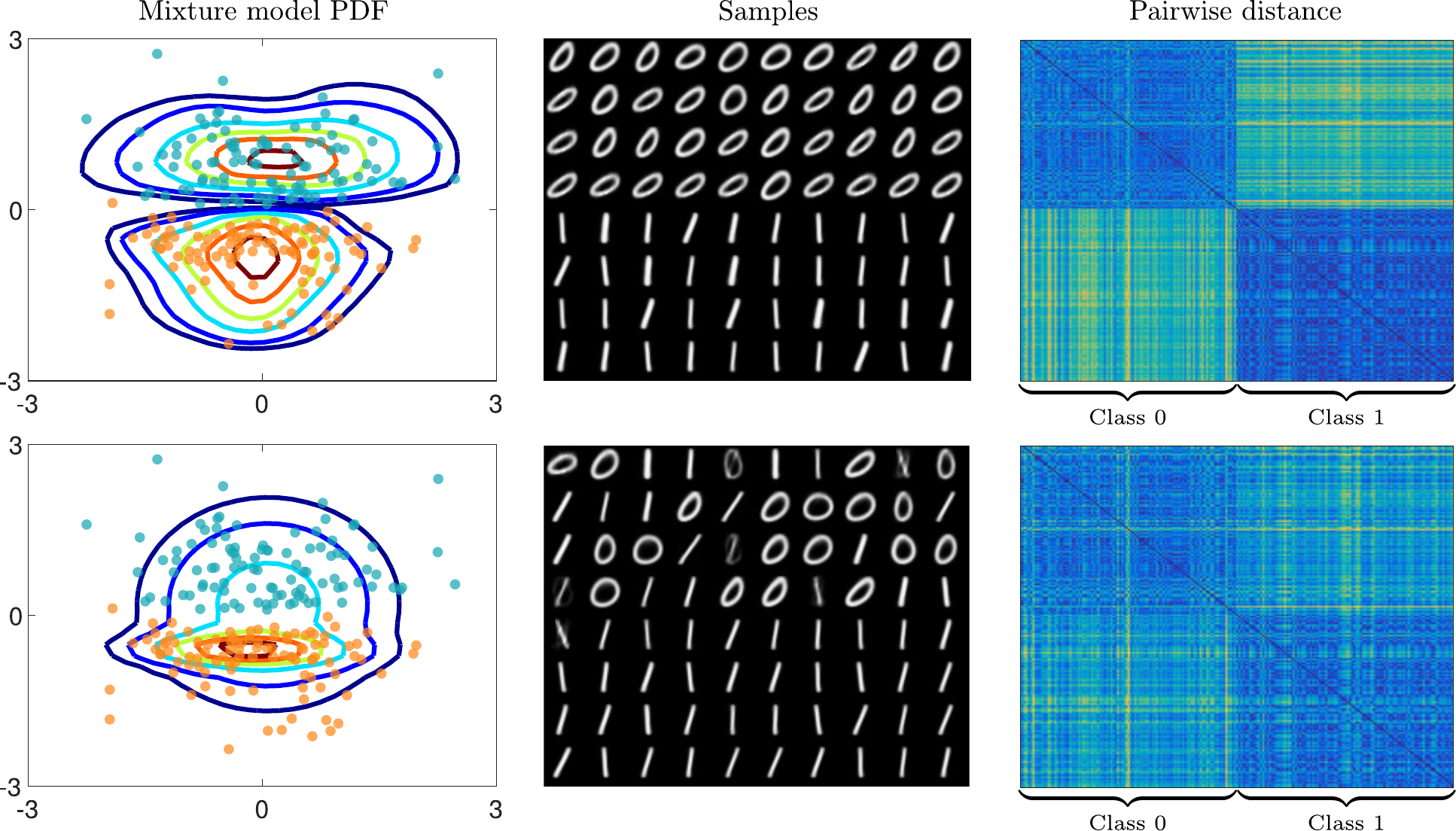}
    \caption{From \textit{left} to \textit{right}: the mixture models, generated samples, and pairwise distances. \textit{Top} row corresponds to the Riemannian model and \textit{bottom} row to the Euclidean model.}
	\label{fig:deep_land_clusters_samples_distances}
  \end{figure}

  We have seen strong indications that the Riemannian metric gives a 
  more meaningful view of the latent space, which may also
  improve probability distributions in the latent space.
  A relevant candidate distribution is the \emph{locally adaptive normal
  distribution (LAND)} \citep{arvanitidis:nips:2016}
  \begin{align}
    \mathrm{LAND}(\b{z}~|~\bs{\mu}, \bs{\Sigma})
      &\propto \exp\left( -\frac{1}{2}\mathrm{dist}_{\bs{\Sigma}}^2 (\b{z}, \bs{\mu}) \right),
  \end{align}
  where $\mathrm{dist}_{\bs{\Sigma}}$ is the Riemannian extension of
  Mahalanobis distance. We fit a mixture of two LANDs to the MNIST data from
  Sec.~\ref{sec:interpolate} alongside a mixture of Euclidean normal distributions.
  The first column of Fig.~\ref{fig:deep_land_clusters_samples_distances} shows
  the density functions of the two mixture models. Only the Riemannian model
  reveals the underlying clusters. We then sample 40 points from each component of
  these generative models\footnote{We do not follow common practice and
    sort samples by their likelihood, as this hides low-quality samples.}
  (center column of the figure). We see that the 
  Riemannian model generates high-quality samples, whereas the Euclidean model
  generates several samples in regions where the generator is not trained
  and therefore produces blurry images. 
  Finally, the right column of Fig.~\ref{fig:deep_land_clusters_samples_distances}
  shows all pairwise distances between the latent points under both Riemannian and
  Euclidean distances. Again, we see that the geometric view clearly reveals
  the underlying clusters.

\subsection{Random Walk on the Data Manifold}\label{ex:random_walk}
\begin{figure}
    \resizebox{\textwidth}{!}{
  	\includegraphics[height=3cm]{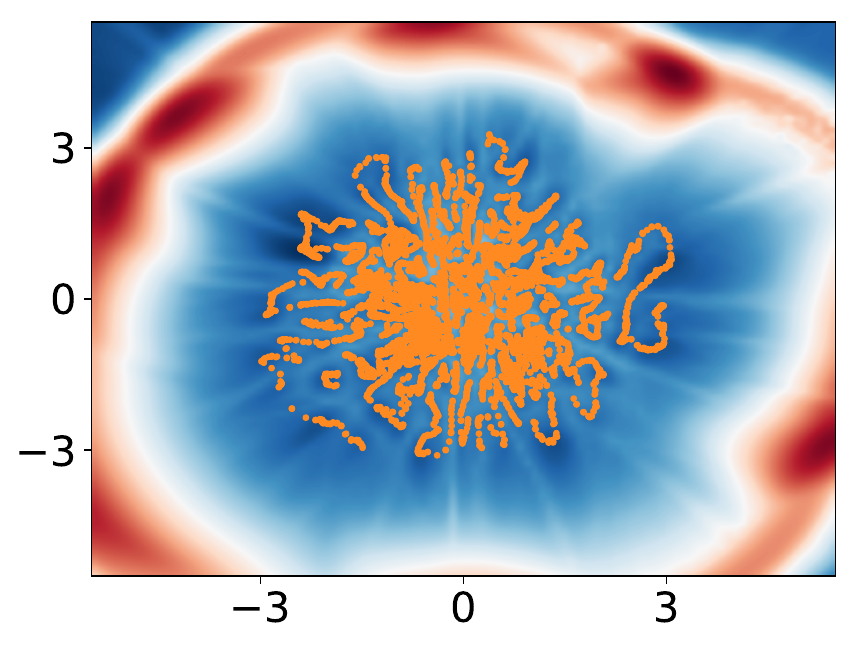}
  	\includegraphics[height=3cm]{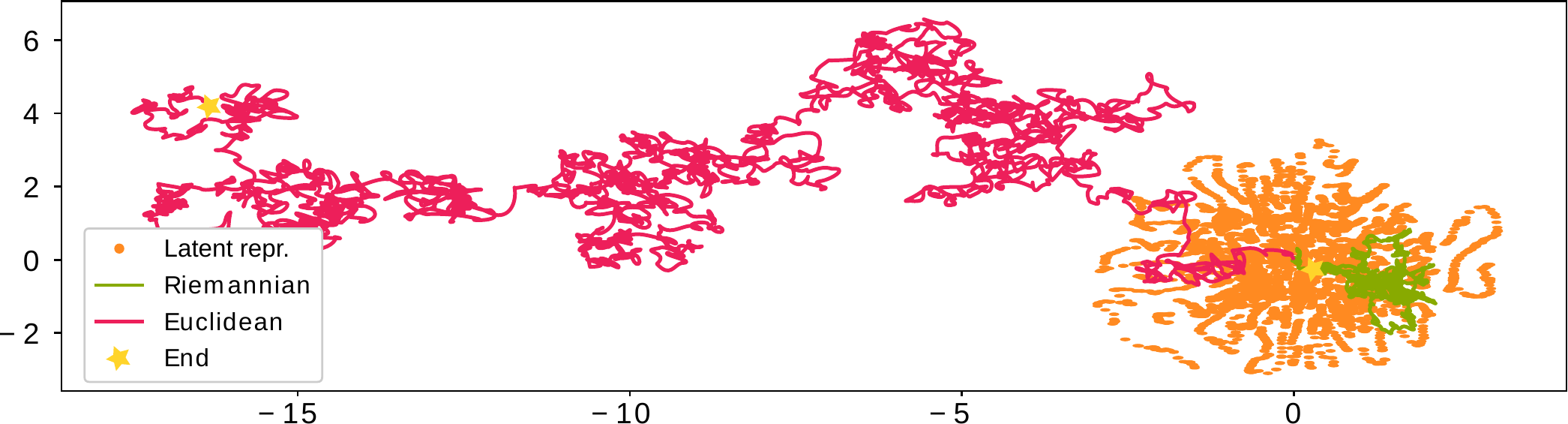}
  	}
    \caption{\textit{Left:} the measure in the latent space. \textit{Right:} the random walks.}
      \label{fig:random_walk}
  \end{figure}

\begin{figure}[h]
	\centering
	\includegraphics[width=1\textwidth]{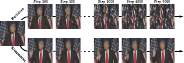}
	\caption{The comparison of the random walks, at the steps 200, 300, 3000, 4000 and 5000.} 
	\label{fig:random_walk_comparison}
  \end{figure}

  Finally, we consider random walks over the data manifold, which is a common tool
  for exploring latent spaces. To avoid the walk drifting outside the data support, practical implementations
  artificially restrict the walk to stay inside the $[-1, 1]^d$ hypercube. Here, we
  consider unrestricted Brownian motion under both the Euclidean and Riemannian
  metric. We perform this random walk in the latent space of the convolutional
  VAE from Sec.~\ref{sec:interpolate}. Figure~\ref{fig:random_walk} shows
  example walks, while Fig.~\ref{fig:random_walk_comparison} shows generated
  images (video \href{https://www.dropbox.com/sh/8ncrsmhaf3x1s80/AAD-JBY-5PcZGQPzXilreSOaa?dl=0}{here}).
  While the Euclidean random walk moves freely, the Riemannian
  walk stays within the support of the data. This is explained in the left panel
  of Fig.~\ref{fig:random_walk}, which shows that the variance term in the
  Riemannian metric creates a ``wall'' around the data, which the random walk will
  only rarely cross. These ``walls'' also force shortest paths to
  follow the data.

\section{Related Work}\label{sec:related}

\paragraph{Generative models.} This unsupervised learning category attracted a lot of attention, especially, due to the advances on the deep neural networks. We have considered VAEs \citep{kingma:iclr:2014, rezende:icml:2014}, but the ideas extend to similar related models. These include extensions that provide more flexible approximate posteriors \citep{rezende:icml:2015, kingma:nips:2016}. 
GANs \citep{goodfellow:nips:2014} also fall in this category, as these models have an explicit generator. While the inference network is not a necessary component in the GAN model, it has been shown that incorporating it improves overall performance \citep{donahue:iclr_2017, dumoulin:iclr_2017}. The same thoughts hold for approaches that transform the latent space through a sequence of bijective functions \citep{dinh:iclr:2017}

\paragraph{Geometry in neural networks.} \citet{bengio:tpami:2013} discuss the importance of geometry in neural networks as a tool to understand local generalization. For instance, the Jacobian matrix is a measure of smoothness for a function that interpolates a surface to the given data. This is exactly the implication in \citep{rifai:icml:2011}, where the norm of the Jacobian acts as a regularizer for the deterministic autoencoder. Recently, \citet{kumar:nips:2017} used the Jacobian to inject invariances in a classifier.

\paragraph{Riemannian Geometry.} 
Like the present paper, \citet{Tosi:UAI:2014} derive a suitable Riemannian
metric in Gaussian process (GP) latent variable models \citep{lawrence2005probabilistic}, but
the computational complexity of GPs causes practical concerns.
Unlike works that explicitly learn a Riemannian metric \citep{hauberg:nips:2012, peltonen:nn:2004},
our metric is fully derived from the generator and requires no extra learning
once the generator is available.

\section{Discussion and Further Extensions}\label{sec:discuss}

The geometric interpretation of representation learning is that the latent space is a compressed and flattened version of the data manifold. 
We show that the \emph{actual} geometry of the data manifold can be more complex than it first appears. 

Here we have initiated the study of proper geometries for generative models. 
We showed that the latent space not only provides a low-dimensional representation of the data manifold, but
at the same time, can reveal the underlying geometrical structure.  
We proposed a new variance network for the generator, which provides meaningful uncertainty estimates while regularizing the geometry.
The new detailed understanding of the geometry provides us with more relevant distance measures, as demonstrated by the fact that
a $k$-means clustering, on these distances, is better aligned with the ground truth label structure than a clustering based on conventional Euclidean distances. 
 We also found that the new distance measure produces smoother interpolation, and when training  Riemannian ``LAND'' mixture models based on the new geometry, the components aligned much better with the ground truth group structure. 
Finally, inspired by the recent  interest in sequence generation by random walks in latent space, we found that geometrically informed random walks stayed on the manifold for much longer runs than sequences based on Euclidean random walks. 


The presented analysis easily extends to sophisticated generative models, where the latent space will be potentially endowed with more flexible nonlinear structures. This directly implies particularly interesting geometrical models. An obvious question is: can the geometry of the latent space play a role  while we learn the generative model? 
Either way, we believe that this geometric perspective provides a new way of thinking and further interpreting the generative models, while at the same time it encourages development of new nonlinear models in the representation space.



\subsubsection*{Acknowledgments} 

LKH is supported by Innovation Fund Denmark / the Danish Center for Big Data Analytics Driven Innovation. SH was supported by a research grant (15334) from VILLUM FONDEN. This project has received funding from the European Research Council (ERC) under the European Union's Horizon 2020 research and innovation programme (grant agreement n\textsuperscript{o} 757360). We gratefully acknowledge the support of the NVIDIA Corporation with the donation of the used Titan Xp GPU. We thank Marcelo Hartmann for initiating the discussion related to the Christoffel symbols.

\clearpage

\bibliography{iclr2018_bibliography}
\bibliographystyle{iclr2018_conference}

\clearpage

\appendix
\section{The Derivation of the Geodesic Differential Equation}\label{sec:geodesic_system_derivation}
The shortest path between two points $\b{x},\b{y}\in\M$ on a Riemannian manifold $\M$ is found by optimizing the functional
\begin{align}
\bs{\gamma}_t^{(\text{shortest})} = \argmin_{\bs{\gamma}_t} \int_0^1 \sqrt{\inner{\dot{\bs{\gamma}}_t}{\b{M}_{\bs{\gamma}_t}\dot{\bs{\gamma}}_t}} \dif{d}t, \quad \bs{\gamma}(0) = \b{x}, ~ \bs{\gamma}(1) = \b{y}
\end{align}
where $\bs{\gamma}_t:[0,1] \rightarrow \M$ and $\dot{\bs{\gamma}}_t = \parder{\bs{\gamma}_t}{t}$. The minima of this problem can be found instead by optimizing the \textit{curve energy}  \citep{docarmo:1992}, so the functional becomes
\begin{align}
\bs{\gamma}_t^{(\text{shortest})} = \argmin_{\bs{\gamma}_t} \frac{1}{2}\int_0^1 {\inner{\dot{\bs{\gamma}}_t}{\b{M}_{\bs{\gamma}_t}\dot{\bs{\gamma}}_t}} \dif{d}t, \quad \bs{\gamma}(0) = \b{x}, ~ \bs{\gamma}(1) = \b{y}.
\end{align}
The inner product can be written explicitly as
\begin{align}
L(\bs{\gamma}_t, \dot{\bs{\gamma}}_t, \b{M}_{\bs{\gamma}_t}) = {\inner{\dot{\bs{\gamma}}_t}{\b{M}_{\bs{\gamma}_t}\dot{\bs{\gamma}}_t}} = \sum_{i=1}^d \sum_{j=1}^d \dot{\gamma}^{(i)}_t \cdot \dot{\gamma}^{(j)}_t \cdot M^{(ij)}_{\bs{\gamma}_t} = \vectorize{\b{M}_{\bs{\gamma}_t}}^\t (\dot{\bs{\gamma}}_t \otimes \dot{\bs{\gamma}}_t )
\end{align}
where the index in the parenthesis represents the corresponding element in the vector or matrix. In the derivation the $\otimes$ is the usual Kronecker product and the $\vectorize{\cdot}$ stacks the column of a matrix into a vector so $\vectorize{\b{M}_{\bs{\gamma}_t}}\in\R^{d^2}$. We find the minimizers by the Euler-Lagrange equation
\begin{align}
\label{eq:euler_lagrange}
\parder{L}{\bs{\gamma}_t} = \parder{}{t}\parder{L}{\dot{\bs{\gamma}}_t}
\end{align}
where
\begin{align}
\parder{}{t}\parder{L}{\dot{\bs{\gamma}}_t}=\parder{}{t} \parder{\inner{\dot{\bs{\gamma}}_t}{\b{M}_{{\bs{\gamma}_t}}\dot{\bs{\gamma}}_t}}{\dot{\bs{\gamma}}_t}= \parder{}{t} \left(2 \cdot  {{\b{M}_{{\bs{\gamma}_t}}}}{{\dot{\bs{\gamma}}_t}} \right)= 2\left[\parder{\b{M}_{{\bs{\gamma}_t}} }{t}{\dot{\bs{\gamma}}_t} +  \b{M}_{{\bs{\gamma}_t}} \ddot{\bs{\gamma}}_t\right].
\end{align}
Since the term
\begin{align}
\parder{\b{M}_{{\bs{\gamma}_t}}}{t} = 
\left[\begin{array}{c c c}
\parder{M^{(11)}_{\bs{\gamma}_t}}{t} & \cdots & \parder{M^{(1D)}_{\bs{\gamma}_t}}{t}\\
\parder{M^{(21)}_{\bs{\gamma}_t}}{t} & \cdots & \parder{M^{(2D)}_{\bs{\gamma}_t}}{t}\\
\vdots & \ddots & \vdots\\
\parder{M^{(D1)}_{\bs{\gamma}}}{t} & \cdots & \parder{M^{(DD)}_{\bs{\gamma}_t}}{t}\\
\end{array}\right]=
\left[\begin{array}{c c c}
\parder{M^{(11)}_{\bs{\gamma}_t}}{\bs{\gamma}_t}^\t \dot{\bs{\gamma}}_t & \cdots & \parder{M^{(1D)}_{\bs{\gamma}_t}}{\bs{\gamma}_t}^\t \dot{\bs{\gamma}}_t\\
\parder{M^{(21)}_{\bs{\gamma}_t}}{\bs{\gamma}_t}^\t \dot{\bs{\gamma}}_t & \cdots & \parder{M^{(2D)}_{\bs{\gamma}}}{\bs{\gamma}_t}^\t \dot{\bs{\gamma}}_t\\
\vdots & \ddots & \vdots\\
\parder{M^{(D1)}_{\bs{\gamma}_t}}{\bs{\gamma}_t}^\t \dot{\bs{\gamma}}_t & \cdots & \parder{M^{(DD)}_{\bs{\gamma}_t}}{\bs{\gamma}_t}^\t \dot{\bs{\gamma}}_t\\
\end{array}\right],
\end{align}
we can write the right hand side of the Eq.~\ref{eq:euler_lagrange} as
\begin{align}
\parder{}{t}\parder{L}{\dot{\bs{\gamma}}_t} = 2\left[(\dot{\bs{\gamma}}_t^\t \otimes \Id_d)\parder{\vectorize{\b{M}_{\bs{\gamma}_t}}}{\bs{\gamma}_t} \dot{\bs{\gamma}}_t + \b{M}_{\bs{\gamma}_t} \ddot{\bs{\gamma}}_t\right].
\end{align}

The left hand side term of the Eq.~\ref{eq:euler_lagrange} is equal to
\begin{align}
\parder{L}{\bs{\gamma}_t} = \parder{}{\bs{\gamma}_t} \left( \vectorize{\b{M}_{\bs{\gamma}_t}}^\t (\dot{\bs{\gamma}}_t \otimes \dot{\bs{\gamma}}_t )\right) =  \parder{\vectorize{\b{M}_{\bs{\gamma}_t}}}{\bs{\gamma}_t}^\t (\dot{\bs{\gamma}}_t \otimes \dot{\bs{\gamma}}_t ).
\end{align}

The final system of $2^{\text{nd}}$ order ordinary differential equations is
\begin{align}
\label{eq:ode:appendix}
\ddot{\bs{\gamma}}_t = -\frac{1}{2}\b{M}_{\bs{\gamma}_t}^{-1}\left[2\cdot (\dot{\bs{\gamma}}_t^\t \otimes \Id_d) \parder{\vectorize{\b{M}_{\bs{\gamma}_t}}}{\bs{\gamma}_t}\dot{\bs{\gamma}}_t - \parder{\vectorize{\b{M}_{\bs{\gamma}_t}}}{\bs{\gamma}_t}^\t (\dot{\bs{\gamma}}_t \otimes \dot{\bs{\gamma}}_t)\right].
\end{align}

Note that $(\dot{\bs{\gamma}}_t^\t \otimes \Id_d) \parder{\text{vec}[\b{M}_{\bs{\gamma}_t}]}{\bs{\gamma}_t}\dot{\bs{\gamma}}_t = \partial_{\bs{\gamma}_t} \b{M}_{\bs{\gamma}_t}  (\dot{\bs{\gamma}}_t \otimes \dot{\bs{\gamma}}_t)$, where $\partial_{\bs{\gamma}_t}\b{M}_{\bs{\gamma}_t}=\left[\partial_{\gamma^{(1)}_t} \b{M}_{\bs{\gamma}_t},\dots, \partial_{\gamma^{(d)}_t} \b{M}_{\bs{\gamma}_t}\right]$ with $\partial_{\gamma^{(j)}_t} \b{M}_{\bs{\gamma}_t}\in\R^{d\times d}$ the partial derivative of $\b{M}_{\bs{\gamma}_t}$ with respect to the $j$-th component of the curve. This fact and the notation $\partial_{\bs{\gamma}_t}\text{vec}[\b{M}_{\bs{\gamma}_t}] = \parder{\text{vec}[\b{M}_{\bs{\gamma}_t}]}{\bs{\gamma}_t}$ allows us to rewrite Eq.~\ref{eq:ode:appendix} as
\begin{align}
\label{eq:ode_new:appendix}
    \ddot{\bs{\gamma}}_t 
      &= -\frac{1}{2}\b{M}_{\bs{\gamma}_t}^{-1}\Big[2\cdot \partial_{\bs{\gamma}_t}\b{M}_{\bs{\gamma}_t} - \partial_{\bs{\gamma}_t}{\vectorize{\b{M}_{\bs{\gamma}_t}}}^\t \Big] (\dot{\bs{\gamma}}_t \otimes \dot{\bs{\gamma}}_t).
\end{align}

The classical geodesic equation in Riemannian geometry \citep{docarmo:1992} is written as
\begin{align}
\label{eq:ode_christoffel}
    \ddot{\gamma}^{(k)}_t = \sum_{i=1}^d\sum_{j=1}^d \Gamma^{k}_{ij} \dot{{\gamma}}^{(i)}_t \dot{{\gamma}}^{(j)}_t,\qquad \Gamma^{k}_{ij} = \sum_{l=1}^d m^{kl}(\partial_i m_{jl} + \partial_j m_{li} - \partial_l m_{ij}),
\end{align}
where $\Gamma^k_{ij}$ are the Christoffel symbols with $m^{ij}={{M}_{\bs{\gamma}_t}^{-1}}^{(ij)}$ the $i,j$ element of the inverse metric and $\partial_k m_{ij} = \partial_{\gamma_t^{(k)}}{M}_{\bs{\gamma}_t}^{(ij)}$ the partial derivative of the metric element $i,j$ with respect to the $k$-th component of the curve $\bs{\gamma}_t$. Note that these Christoffel symbols are symmetric i.e. $\Gamma^k_{ij} = \Gamma^k_{ji}$.

Even if the derived ODE system in Eq.~\ref{eq:ode_new:appendix} is equivalent to  Eq.~\ref{eq:ode_christoffel}, the corresponding coefficients are not equal. In particular, let the coefficient matrix $\b{C} = \b{M}_{\bs{\gamma}_t}^{-1}\Big[2\cdot\partial_{\bs{\gamma}_t}\b{M}_{\bs{\gamma}_t} - \partial_{\bs{\gamma}_t}{\vectorize{\b{M}_{\bs{\gamma}_t}}}^\t \Big]\in\R^{d\times d^2}$. We can easily see that $C^{(ij)}\neq C^{(ji)}$, however, comparing Eq.~\ref{eq:ode_new:appendix} with Eq.~\ref{eq:ode_christoffel} shows that due to symmetries the following relation holds between our coefficients $\b{C}$ and the true Christoffel symbols
\begin{align}
    \Gamma_{ij}^k = \Gamma_{ji}^k = \frac{C^{(k p)} + C^{(kq})}{2}, \quad \text{with}\quad p= d(i-1)+j \quad \text{and} \quad q=d(j-1)+i.
\end{align}

\section{The Derivation of the Riemannian Metric}\label{sec:expectation_metic}
The proof of Theorem~\ref{thm:expected_metric}.
  \begin{proof}
As we introduced in Eq.~\ref{eq:stoch_gen} the stochastic generator is
\begin{align}
    f(\b{z}) = \bs{\mu}(\b{z}) + \bs{\sigma}(\b{z}) \odot \bs{\epsilon},
    \qquad
    \bs{\mu}: \Z \to \X,\enspace
    \bs{\sigma}: \Z \to \R_{+}^D,\enspace
    \bs{\epsilon} \sim \N(\b{0},\Id_D).
\end{align}
Thus, we can compute the corresponding Jacobian as follows
\begin{align}
\parder{f(\b{z})}{\b{z}} = \b{J}_{\b{z}} &= \left[ 
\begin{array}{c c c c}
\parder{f^{(1)}_{\b{z}}}{z_1} & \parder{f^{(1)}_{\b{z}}}{z_2} & \cdots & \parder{f^{(1)}_{\b{z}}}{z_d} \\
\parder{f^{(2)}_{\b{z}}}{z_1} & \parder{f^{(2)}_{\b{z}}}{z_2} & \cdots & \parder{f^{(2)}_{\b{z}}}{z_d} \\
\vdots & \vdots & \ddots & \vdots \\
\parder{f^{(D)}_{\b{z}}}{z_1} & \parder{f^{(D)}_{\b{z}}}{z_2} & \cdots & \parder{f^{(D)}_{\b{z}}}{z_d}
\end{array}\right]_{D\times d} \\
&= \underbrace{\left[ 
\begin{array}{c c c c}
\parder{\bs{\mu}^{(1)}_{\b{z}}}{z_1} & \parder{\bs{\mu}^{(1)}_{\b{z}}}{z_2} & \cdots & \parder{\bs{\mu}^{(1)}_{\b{z}}}{z_d} \\
\parder{\bs{\mu}^{(2)}_{\b{z}}}{z_1} & \parder{\bs{\mu}^{(2)}_{\b{z}}}{z_2} & \cdots & \parder{\bs{\mu}^{(2)}_{\b{z}}}{z_d} \\
\vdots & \vdots & \ddots & \vdots \\
\parder{\bs{\mu}^{(D)}_{\b{z}}}{z_1} & \parder{\bs{\mu}^{(D)}_{\b{z}}}{z_2} & \cdots & \parder{\bs{\mu}^{(D)}_{\b{z}}}{z_d}
\end{array}\right]_{D\times d}}_{\b{A}}
+
\underbrace{\left[\b{S}_1 \bs{\epsilon}, ~\b{S}_2 \bs{\epsilon}, ~\cdots, ~\b{S}_d \bs{\epsilon}\right]_{D\times d}}_{\b{B}},\\
\text{where} \quad  \b{S}_i &= 
\left[
\begin{array}{c c c c}
\parder{\sigma^{(1)}_{\b{z}}}{z_i} & 0 & \cdots & 0 \\
0 & \parder{\sigma^{(2)}_{\b{z}}}{z_i} & \cdots & 0\\
\vdots & \vdots & \ddots & \vdots \\
0 & 0 & \cdots & \parder{\sigma^{(D)}_{\b{z}}}{z_i}
\end{array}\right]_{D\times D}, \quad i=1,\dots,d
\end{align}
and the resulting ``random'' metric in the latent space is $\b{M}_{\b{z}} = \b{J}_{\b{z}}^\t \b{J}_{\b{z}}$. The randomness is due to the random variable $\bs{\epsilon}$, and thus, we can compute the expectation 
\begin{align}
\overline{\b{M}}_{\b{z}} = \E_{p(\bs{\epsilon})}[\b{M}_{\b{z}}] 
= \E_{p(\bs{\epsilon})}[ (\b{A} + \b{B})^\t (\b{A} + \b{B})] = \E_{p(\bs{\epsilon})}[ \b{A}^\t \b{A} + \b{A}^\t \b{B} + \b{B}^\t \b{A} + \b{B}^\t \b{B}].
\end{align}

Using the linearity of expectation we get that
\begin{align}
\E_{p(\bs{\epsilon})}[\b{A}^\t \b{B}] = \E_{p(\bs{\epsilon})}\left[\b{A}^\t [\b{S}_1 \bs{\epsilon}, \b{S}_2 \bs{\epsilon}, \cdots, \b{S}_d \bs{\epsilon}]\right] = \b{A}^\t [\b{S}_1 \cancelto{\b{0}}{\E_{p(\bs{\epsilon})}[\bs{\epsilon}]},\dots,\b{0}] = 0
\end{align}
because $\E_{p(\bs{\epsilon})}[\bs{\epsilon}] = \b{0}$. The other term
\begin{align}
 \E_{p(\bs{\epsilon})}[\b{B}^\t \b{B}] &=
\E_{p(\bs{\epsilon})} \left( \left[
\begin{array}{c}
\bs{\epsilon}^\t \b{S}_1 \\
\bs{\epsilon}^\t \b{S}_2\\
\vdots \\
\bs{\epsilon}^\t \b{S}_d
\end{array}
\right]_{d\times D}
\left[ \b{S}_1 \bs{\epsilon}, \b{S}_2 \bs{\epsilon}, \cdots,  \b{S}_d\bs{\epsilon}\right] \right)\\
&=\E_{p(\bs{\epsilon})}\left(
\left[
\begin{array}{c c c c}
\bs{\epsilon}^\t \b{S}_1 \b{S}_1 \bs{\epsilon} & \bs{\epsilon}^\t \b{S}_1 \b{S}_2 \bs{\epsilon} & \cdots & \bs{\epsilon}^\t \b{S}_1 \b{S}_d \bs{\epsilon}\\
\bs{\epsilon}^\t \b{S}_2 \b{S}_1 \bs{\epsilon} & \bs{\epsilon}^\t \b{S}_2 \b{S}_2 \bs{\epsilon} & \cdots & \bs{\epsilon}^\t \b{S}_2 \b{S}_d \bs{\epsilon}\\
\vdots & \vdots & \vdots & \vdots \\
\bs{\epsilon}^\t \b{S}_d \b{S}_1 \bs{\epsilon} & \bs{\epsilon}^\t \b{S}_d \b{S}_2 \bs{\epsilon} & \cdots & \bs{\epsilon}^\t \b{S}_d \b{S}_d \bs{\epsilon}
\end{array}
\right] \right)\\
\text{with}\quad  \E_{p(\bs{\epsilon})}\left[ \bs{\epsilon}^\t \b{S}_i \b{S}_j \bs{\epsilon} \right] &= \E_{p(\bs{\epsilon})}\left[ \left(\epsilon_1 \parder{\sigma^{(1)}_{\b{z}}}{z_i}, \epsilon_2 \parder{\sigma^{(2)}_{\b{z}}}{z_i}, \cdots, \epsilon_D \parder{\sigma^{(D)}_{\b{z}}}{z_i}\right)
\left(\begin{array}{c}
\epsilon_1 \parder{\sigma^{(1)}_{\b{z}}}{z_j}\\
\epsilon_2 \parder{\sigma^{(2)}_{\b{z}}}{z_j}\\
\vdots\\
\epsilon_D \parder{\sigma^{(D)}_{\b{z}}}{z_j}
\end{array}\right)
\right]\\
=\E_{p(\bs{\epsilon})} & \left[
\epsilon_1^2 \left( \parder{\sigma^{(1)}_{\b{z}}}{z_i} \parder{\sigma^{(1)}_{\b{z}}}{z_j} \right) + \epsilon_2^2 \left( \parder{\sigma^{(2)}_{\b{z}}}{z_i} \parder{\sigma^{(2)}_{\b{z}}}{z_j} \right) + \cdots \epsilon_D^2 \left( \parder{\sigma^{(D)}_{\b{z}}}{z_i} \parder{\sigma^{(D)}_{\b{z}}}{z_j}  \right)
\right]\\
&=diag(\b{S}_i)^\t diag(\b{S}_j),
\end{align}
because $\E_{p(\bs{\epsilon})}[\epsilon_i^2] = 1,~ \forall i=1,\dots,D$.

The matrix $\b{A} = \b{J}_{\b{z}}^{(\bs{\mu})}$ and for the variance network
\begin{align}
\b{J}_{\b{z}}^{(\bs{\sigma})} = \left[ 
\begin{array}{c c c c}
\parder{\bs{\sigma}^{(1)}_{\b{z}}}{z_1} & \parder{\bs{\sigma}^{(1)}_{\b{z}}}{z_2} & \cdots & \parder{\bs{\sigma}^{(1)}_{\b{z}}}{z_d} \\
\parder{\bs{\sigma}^{(2)}_{\b{z}}}{z_1} & \parder{\bs{\sigma}^{(2)}_{\b{z}}}{z_2} & \cdots & \parder{\bs{\sigma}^{(2)}_{\b{z}}}{z_d} \\
\vdots & \vdots & \ddots & \vdots \\
\parder{\bs{\sigma}^{(D)}_{\b{z}}}{z_1} & \parder{\bs{\sigma}^{(D)}_{\b{z}}}{z_2} & \cdots & \parder{\bs{\sigma}^{(D)}_{\b{z}}}{z_d}
\end{array}\right]
\end{align}
it is easy to see that  $\E_{p(\bs{\epsilon})}[\b{B}^\t \b{B}] = \left(\b{J}_{\b{z}}^{(\bs{\sigma})}\right)^\t \b{J}_{\b{z}}^{(\bs{\sigma})}$. So the expectation of the induced Riemannian metric in the latent space by the generator is
\begin{align}
\bar{\b{M}}_{\b{z}} = \left(\b{J}_{\b{z}}^{(\bs{\mu})}\right)^\t \b{J}_{\b{z}}^{(\bs{\mu})} + \left(\b{J}_{\b{z}}^{(\bs{\sigma})}\right)^\t \b{J}_{\b{z}}^{(\bs{\sigma})}
\end{align}
which concludes the proof.
\end{proof}

\section{Influence of Variance on the Marginal Likelihood}\label{sec:marginal_likelihood_modeling}

We trained a VAE on the digits 0 and 1 of the MNIST scaled to $[-1,1]$. We randomly split the data to $90\%$ training and $10\%$ test data, ensuring balanced classes. First, we only trained the encoder and the mean function of the decoder. Then, keeping these fixed, we trained two variance functions: one based on standard deep neural network architecture, and the other using our proposed RBF model. Clearly, we have two generators with the same mean function, but different variance functions. Below we present the architectures for the standard neural networks. For the RBF model we used 32 centers and $a=1$.

\begin{center}
  \begin{tabular}{ c c c c }
Encoder/Decoder & Layer 1 & Layer 2 & Layer 3\\\hline\hline
 $\bs{\mu}_{\phi}$ & 64, (\textit{softplus}) & 32, (\textit{softplus}) & $d$, (\textit{linear})\\
 $\bs{\sigma}_{\phi}$ & 64, (\textit{softplus}) & 32, (\textit{softplus}) & $d$, (\textit{softplus})\\\hline
  $\bs{\mu}_{\theta}$ & 32, (\textit{softplus}) & 64, (\textit{softplus}) & $D$, (\textit{tanh})\\
    $\bs{\sigma}_{\theta}$ & 32, (\textit{softplus}) & 64, (\textit{softplus}) & $D$, (\textit{softplus})\\
    \hline
  \end{tabular}
\end{center}

The numbers corresponds to the layer size together with the activation function in parenthesis. Further, the mean and the variance functions share the weights of the first layer. The input space dimension is $D={784}$. Then, we computed the marginal likelihood $p(\b{x})$ of the test data using Monte Carlo as:

\begin{align}
p(\b{x}) = \int_{\Z} p(\b{x} | \b{z}) p(\b{z}) \dif{d}\b{z} \simeq \frac{1}{S} \sum_{s=1} p(\b{x} | \b{z}_s), \quad \b{z}_s \sim p(\b{z})
\end{align}
using $S=10000$ samples. The generator with the standard variance function achieved -68.25 mean log-marginal likelihood, while our proposed model -50.34, where the higher the better.


The reason why the proposed RBF model performs better can be easily analyzed. The marginal likelihood under the Monte Carlo estimation is, essentially, a large Gaussian mixture model with equal weights $\frac{1}{S}$. Each mixture component is defined by the generator through the likelihood $p(\b{x}|\b{z})= \N\left(\b{x}~|~\bs{\mu}_{\theta}(\b{z}), \Id_D \bs{\sigma}^2_{\theta}(\b{z})\right)$. Considering the variance term, the standard neural network approach is trained on the given data points and the corresponding latent codes. Unfortunately, its behavior is arbitrary in regions where there are not any encoded data. On the other hand our proposed model assigns large variance to these regions, while on the regions where we have latent codes its behavior will be approximately the same with the standard neural network.
This implies that the resulting marginal likelihood $p(\b{x})$ for the two models are highly similar in regions of high data density, but significantly different elsewhere. The RBF variance model ensures that mixture components in these regions have high variance, whereas the standard architecture assign arbitrary variance. Consequently, the RBF-based $p(\b{x})$ assigns minimal density to regions with no data, and, thus, attains higher marginal likelihood elsewhere.


\section{Implementation Details for the Experiments}\label{sec:model_details}

\begin{algorithm}[h]
	\caption{The training of a VAE that ensures geometry}
	\label{alg:mle}
      \algsetup{indent=1em}
        \begin{algorithmic}[1]
         	\ENSURE {the estimated parameters of the neural networks $\theta, \phi, \psi$}
			 	\STATE{Train the $\bs{\mu}_{\phi}, \bs{\sigma}_{\phi}, \bs{\mu}_{\theta}$ as in \citet{kingma:iclr:2014}, keeping $\bs{\sigma}_{\psi}$ fixed.}
			 				 	\STATE{Train the $\bs{\sigma}_{\psi}$ as explained in Sec.~\ref{sec:proper_geometry}.}
        \end{algorithmic}
\end{algorithm}

\paragraph{Details for Experiments \ref{ex:algorithm_extension}, \ref{sec:interpolate} \& \ref{ex:land}.}

The pixel values of the images are scaled to the interval $[0,1]$. We use for the functions $\bs{\mu}_{\phi}, \bs{\sigma}_{\phi}, \bs{\mu}_{\theta}$ multilayer perceptron (MLP) deep neural networks, and for the $\bs{\beta}_{\psi}$ the proposed RBF model with 64 centers, so $\b{W} \in \R^{D \times 64}$ and the parameter $a$ of Eq.~\ref{eq:rbf_bandwidth} is set to 2. We used $L_2$ regularization with parameter equal to $1e^{-5}$.

\begin{center}
  \begin{tabular}{ c c c c }
Encoder/Decoder & Layer 1 & Layer 2 & Layer 3\\\hline\hline
 $\bs{\mu}_{\phi}$ & 64, (\textit{tanh}) & 32, (\textit{tanh}) & $d$, (\textit{linear})\\
 $\bs{\sigma}_{\phi}$ & 64, (\textit{tanh}) & 32, (\textit{tanh}) & $d$, (\textit{softplus})\\\hline
  $\bs{\mu}_{\theta}$ & 32, (\textit{tanh}) & 64, (\textit{tanh}) & $D$, (\textit{sigmoid})\\
    \hline
  \end{tabular}
\end{center}

The number corresponds to the size of the layer, and in the parenthesis the activation function.
For the encoder, the mean and the variance functions share the weights of the Layer 1. The input space dimension $D={784}$. After the training, the geodesics can be computed by solving Eq.~\ref{eq:ode} numerically. The LAND mixture model is fitted as explained in \citep{arvanitidis:nips:2016}.

\paragraph{Details for Experiments \ref{ex:random_walk}.}

In this experiment we used Convolutional Variational Auto-Encoders. The pixel values of the images are scaled to the interval $[0,1]$. For the $\bs{\beta}_{\psi}$ we used the proposed RBF model with 64 centers and the parameter $a$ of Eq.~\ref{eq:rbf_bandwidth} is set to 2.

Considering the variance network during the decoding stage, the RBF generates an image, which represents intuitively the total variance of each pixel for the decoded final image, but in an initial sub-sampled version. Afterwards, this image is passed through a sequence of deconvolution layers, and at the end will represent the variance of every pixel for each RGB channel. However, it is critical that the weights of the filters must be clipped during the training to $\R_+$ to ensure positive variance.

\begin{center}
  \begin{tabular}{ c c c c c}
 Encoder & Layer 1 (Conv) & Layer 2 (Conv) & Layer 3 (MLP) & Layer 4 (MLP)\\\hline\hline
 $\bs{\mu}_{\phi}$ & 32, 3, 2, (\textit{tanh}) & 32, 3, 2, (\textit{tanh}) & 1024, (\textit{tanh}) & $d$, (\textit{linear})\\
  $\bs{\sigma}_{\phi}$ & 32, 3, 2, (\textit{tanh}) & 32, 3, 2, (\textit{tanh}) & 1024, (\textit{tanh}) & $d$, (\textit{softplus})\\
    \hline
  \end{tabular}
\end{center}

For the convolutional and deconvolutional layers, the first number is the number of applied filters, the second is the kernel size, and third is the stride. Also, for the encoder, the mean and the variance functions share the convolutional layers. We used $L_2$ regularization with parameter equal to $1e^{-5}$.

\begin{center}
  \begin{tabular}{ c c c c c c c}
 Decoder & L. 1 (MLP) & L. 2 (MLP) & L. 3 (DE) & L. 4 (DE) & L. 5 (DE) & L.6 (CO) \\\hline\hline
 $\bs{\mu}_{\theta}$ & 1024, (\textit{t}) & $D/4$, (\textit{t}) & 32, 3, 2, (\textit{t}) & 32, 3, 2, (\textit{t}) & 3, 3, 1, (\textit{t}) & 3, 3, 1, (\textit{s}) \\
    \hline
  \end{tabular}
\end{center}

For the decoder, the acronyms (DE) = Deconvolution, (CO) = Convolution and $(t)$, $(s)$ stand for \textit{tanh} and \textit{sigmoid}, respectively. Also, $D=width \times height \times channels$ of the images, in our case 64,64,3. For all the convolutions and deconvolutions, the padding is set to \textit{same}. We used $L_2$ regularization with parameter equal to $1e^{-5}$.

\begin{center}
  \begin{tabular}{ c c c c}
 Decoder & Layer 1 (RBF) & Layer 2 (Deconv) &  Layer 3 (Conv)\\\hline\hline
 $\bs{\beta}_{\psi}$ & $W\in\R^{(D/2)\times 64}$&  1, 3, 2 (\textit{linear}) & 3, 3, 1 (\textit{linear})\\
    \hline
  \end{tabular}
\end{center}

The Brownian motion over the Riemannian manifold in the latent space is presented in Alg.~\ref{alg:brownian_motion}.

\begin{algorithm}[h]
	\caption{Brownian motion on a Riemannian manifold}
	\label{alg:brownian_motion}
      \algsetup{indent=1em}
        \begin{algorithmic}[1]
         	\REQUIRE{the starting point $\b{z}\in\R^{d\times 1}$, stepsize $s$, number of steps $N_s$, the metric tensor $\b{M}(\cdot)$.}
         	\ENSURE {the random steps $\b{Z}\in\R^{N_s \times d}$.}
         	\FOR{$n=0$ to $N_s$}
		\STATE $\b{L}, \b{U} = eig\left(\b{M}(\b{z})\right)$,$\qquad$ ($\b{L}$: eigenvalues, $\b{U}$: eigenvectors) 
		\STATE $\b{v} = \b{U} \b{L}^{-\frac{1}{2}} \bs{\epsilon}$,$\qquad$ $\bs{\epsilon}\sim \N(\b{0}, \Id_d)$
		\STATE $\b{z} = \b{z} + s \cdot \b{v}$
		\STATE $\b{Z}(n,~:)=\b{z}$
		\ENDFOR
        \end{algorithmic}
\end{algorithm}

\end{document}